\documentclass[12pt,oneside]{article}

 \usepackage{latexsym}
 \usepackage{mathrsfs}
 \usepackage{amsfonts}
 \usepackage{amssymb}
 \usepackage{epsfig}       
 \usepackage{subfigure}    
 \usepackage{graphicx}
 \usepackage{float}
 \usepackage{multirow}
 \usepackage{array}
 \usepackage{xcolor}

 \usepackage{epstopdf}

 \usepackage{palatino}
 \usepackage{amsmath}
 \usepackage{tikz}
 \usetikzlibrary{shapes.geometric, arrows}

 \def\reff#1{{\rm(\ref{#1})}}

\def\argmin{{\rm argmin}\,}

\begin{document}

 \newtheorem{definition}{Definition}[section]
 \newtheorem{lemma}{Lemma}[section]
 \newtheorem{corollary}{Corollary}[section]
 \newtheorem{theorem}{Theorem}[section]
 \newtheorem{remark}{Remark}[section]
 \newtheorem{example}{Example}[section]
 \newtheorem{assumption}{Assumption}[section]
 \newtheorem{property}{Property}[section]
 \newtheorem{proposition}{Proposition}[section]

 \title{A Novel Euler's Elastica based Segmentation Approach for Noisy Images via using the Progressive Hedging Algorithm
}

 \author{Lu Tan\thanks{School of EECMS, Curtin University, Perth, Australia. Email: lu.tan1@postgrad.curtin.edu.au}, Ling Li\thanks{School of EECMS, Curtin University, Perth, Australia. Email: l.li@curtin.edu.au}, Wanquan Liu\thanks{School of EECMS, Curtin University, Perth, Australia. Email: w.liu@curtin.edu.au}, Jie Sun\thanks{School of EECMS, Curtin University, Perth, Australia. Email: jie.sun@curtin.edu.au}, Min Zhang \thanks{School of EECMS, Curtin University, Perth, Australia. Email: min.zhang2@curtin.edu.au}
 }

\maketitle

\begin{abstract}
 \noindent Euler's elastica based unsupervised segmentation models have strong capability of completing the missing boundaries for existing objects in a clean image, but they are not working well for noisy images. This paper aims to establish a Euler's elastica based approach that properly deals with the random noises to improve the segmentation performance for noisy images. We solve the corresponding optimization problem via using the progressive hedging algorithm (PHA) with a step length suggested by the alternating direction method of multipliers (ADMM). Technically, all the simplified convex versions of the sub problems derived from the major framework of PHA can be obtained by using the curvature weighted approach and the convex relaxation method. Then an alternating optimization strategy is applied with the merits of using some powerful accelerating techniques including the fast Fourier transform (FFT) and generalized soft threshold formulas. Extensive experiments have been conducted on both synthetic and real images, which validated some significant gains of the proposed segmentation models and demonstrated the advantages of the developed algorithm.
 \vspace{5mm}

 \noindent {\bf Keywords}\hspace{2mm} Euler's elastic energy, stochastic noises, progressive hedging algorithm (PHA), alternating direction method of multipliers (ADMM), curvature weighted approach
 \vspace{2mm}
%

\end{abstract}

\section{Introduction}

In image segmentation, the Mumford-Shah model proposed by Mumford D and Shah J \cite{ms1989} is regarded as the most significant region-based model and has been applied to  many applications. In 2001, the two-phase Chan-Vese (CV) model \cite{ch2001} was proposed   to detect objects in a given image. With the increasing complexity of images, the multiphase segmentation models \cite{f2010,ywf2012,brd2010} were proposed and these models mainly represent different regions by using the level set functions \cite{wyg2018,tpldww2017,zchw2006,tll2018}. In order to reduce the number of level set functions, Chan et al. proposed a multiphase segmentation model \cite{vch2002}, which is a generalization of CV model.

Some specific segmentation models \cite{pd2002,mrgg2004,s2013} were also established subsequently according to different noise distribution. They obtained the characteristic information contained in images by estimating corresponding parameters. When dealing noisy images, it is known that many segmentation problems need a suitable noise model, e.g., synthetic aperture radar, positron emission tomography, electron micrograph or medical ultrasound imaging, etc. Especially when the data were collected with poor statistics, it is necessary to consider the influence of the noise probability distribution in segmentation implementation.

Recently, authors in \cite{ztch2013,td2017,tpldww2017} made some progress in achieving illusory contour recovery while doing segmentation, which can identify absent boundaries or missing shapes successfully without necessary region features. In detail, \cite{ztch2013} employed the fitting terms of two-phase CV model and the Euler's elastica term \cite{td2017} as the regularization. Its major contribution was that the missing boundaries were interpolated automatically without specifying the regions. \cite{tpldww2017} improved the segmentation with depth problem \cite{nms1993,zchw2006} and achieved acceleration via the strategies of model simplification and constraint projection. Its significant performance enhancements included shape reconstruction of occluded objects and determination of their ordering relation in a specific scene based on only one single image. Many other works illustrated that the curvature-related terms have played crucial roles in the boundary reconstruction \cite{kzs2014,tll2018} and image restoration \cite{ztch2013-1,tlp2018} with the capacity of producing excellent edge and corner preservation results. All of these researches show the significant potential for curvature-based methods.

However, the current segmentation models mentioned above cannot be directly applied for noisy images when the type of noise is unknown or there are more than one type of noise in the image. The reason is that in these models there exists a one-to-one mapping relationship between the parameters to be evaluated and the noisy images with some probability density distribution. Besides, the curvature-related terms will bring extra computational complexity due to the existence of nonlinear higher-order derivatives. This issue also appears in other variational models such as the non-texture image inpainting \cite{yk2016} and image denoising \cite{ynl2016} with features (edge, corner, smoothness, contrast, etc.) preservation. Hence it is essential to take some mathematical optimization techniques, for example with global solution, stability guarantee and calculation acceleration, into consideration in the process of algorithm design.

For the difficulty of dealing with stochastic noises which are inherently generated by the acquisition procedure of imaging due to various issues, \cite{rw2017,rs2018} have given us a great source of inspiration. The authors extended the range of applications for progressive hedging algorithm (PHA) in multistage stochastic variational inequality problems and explored stochastic complementarity problems in a two-stage formulation. Convergence of the algorithm was proved in detail and how the PHA performs was validated through numerical experiments as well as its stability and practicability. One of our major motivations is to embed stochastic property into segmentation energy functional in images with unknown noises or arbitrary damages and then implement PHA to solve it. Furthermore, Euler's elastica term will be also considered as the regularization in our variational formulations design since its better properties in dealing with image feature information. Last but not least, in order to improve the computational efficiency and solve problems caused by the non-convex, non-differentiable, nonlinear and higher order terms involved in Euler's elastica related functional, fast algorithms of alternating direction method of multipliers (ADMM) \cite{tpldww2017,tll2018,tlp2018,gpt2016} and curvature weighted approach \cite{yk2016,bst2011,dgt2018} will be systematically designed in the PHA algorithm framework as a fusion for energy minimization problems. The creative introduction of Euler's elastica term along with the nontrivial task of its analytical study will be another main goal of our research.

Our contributions in this paper can be summarized in the following aspects:
\begin{itemize}
\item[(i)] We intend to propose novel formulations for image segmentation with stochastic noises for various applications by transforming the original minimization problems into the optimization framework of stochastic programming.
\item[(ii)] Euler's elastica term is employed to realize completion of meaningful missing boundaries and reconstruction of occluded structures of objects, which further enhances the segmentation performance.
\item[(iii)] Our novel variational formulations will be applied in the problems of two-phase Euler's elastica based segmentation and segmentation with depth in gray and color spaces respectively.
\item[(iv)] A general numerical algorithm based on PHA is proposed to calculate the novel formulations. Fusion of ADMM and curvature weighted approach (ADMM-C) is designed for the minimization of the Euler's elastica energy related sub variational problems. The minimization problems derived from ADMM-C will be then efficiently solved by Fast Fourier transform (FFT) \cite{dgt2018,mps2014} and analytical soft threshold formulas \cite{tll2018,tlp2018}.
\end{itemize}

The rest of this paper is structured as follows. Section 2 briefly reviews the  related approaches in this field. Our proposed approach and algorithm framework are presented in Section 3. The experiments conducted with performance evaluation and comparison are described in Section 4 followed by the conclusion in Section 5.

\section{Research background}

For the purpose of clarifying the motivations in this paper, some related works will be briefly reviewed before presenting our contributions clearly in Sections 3.

\subsection{Euler's elastica based segmentation}

Illusory contour capture and shape reconstruction \cite{k2017,pg2017} is a challenging problem which aims to complete the missing boundaries or fuzzy areas for existing objects in an image. It is a very common phenomenon in human vision. Part of illusory contours consists of objects' actual boundaries and the other part is made of missing perceptual edges. However, current computer technique can only deal with closed boundaries. It is extremely difficult for computers to identify illusory contours automatically. Euler's elastica based segmentation techniques have one place in this problem due to their crucial roles in boundary reconstruction and image restoration.

\textbf{Two-phase Euler's elastica based segmentation:} Zhu, Tai, and Chan \cite{ztch2013}  proposed the Chan-Vese-Euler (CVE) model designed for the foreground shape recovery problem by combining the CV model \cite{ch2001} and Euler elastica regularizer \cite{td2017}. This model could recover the illusory contours and form a complete meaningful object, even without requiring initialization of fixed points. According to their work, the energy functional is defined as
\begin{eqnarray}\label{1}
E(\phi,c)=\alpha_1 \int_{\Omega} (f-c_1)^2\phi dx + \alpha_2 \int_{\Omega} (f-c_2)^2(1-\phi) dx + \int_{\Omega} (\alpha+\beta\kappa^2)|\nabla\phi| dx,
\end{eqnarray}
where $\mu, \alpha, \beta$ are positive penalty parameters, $\phi$ is a binary level set representation supposed to take on either 0 or 1. The last term of this functional is the classic Euler's elastica term. $\kappa$ denotes the curvature represented as $\kappa=\nabla(\nabla\phi/|\nabla\phi|)$.

\textbf{Segmentation with depth information:} In \cite{nms1993}, Nitzberg, Mumford and Shiota defined the problem of segmentation with depth information as a problem of recovering occluded shapes and their ordering relations based on a 2D image. The variables defined in this problem are in three folds: 1) the shapes of the regions $R_1, R_2, \cdots, R_n$ to which different objects belong; 2) the ordering relations among objects; 3) the pixel intensities of objects. Without loss of generality, one can assume that the objects $R_1, R_2, \ldots, R_n$ in an image are in ascending order, i. e., $R_1$ is the nearest object to the observer while $R_n$ is the farthest one (i.e. background). Let $R_i'$ be defined as the visible part of $R_i$, i.e., $R_1'=R_1$, $R_i'=R_i-\Box_{j<i} R_j$, ($i=2, \ldots, n$). In addition, $R_{n+1}'=\Omega-\Box_{j<n+1}R_j$ is defined as the visible background. Based on the above assumptions and definitions, the level set based energy functional is formulated as in \cite{zchw2006}
\begin{eqnarray}\label{2}
E(\varphi,c)&=&\sum_{i=1}^n \int_{\Omega} (\alpha+\beta|\kappa_i|)|\nabla\varphi_i|\delta(\varphi_i) dx + \int_{\Omega} (f-c_{n+1})^2 \prod_{j=1}^{n} (1-H(\varphi_j)) dx  \nonumber\\
&& + \sum_{i=1}^n\left( \int_{\Omega} (f-c_i)^2 H(\varphi_i) \prod_{j=1}^{i-1} (1-H(\varphi_j)) dx \right),
\end{eqnarray}
where $\alpha, \beta$ are two positive penalty parameters, $c_i\in R_i$ is the pixel intensity of the $i$-th object, and $f$ is the image to be processed. $\kappa_i$ denotes the curvature of boundary for region $R_i$. Here $|\kappa|$ is chosen to substitute the square in Euler's elastica term with the reason that the the object corners can be preserved when $|\kappa|$ becomes large. The level set function $\varphi$ is represented by a continuous signed distance function. $H(x)$ and $\delta(x)$ are Heaviside function and Dirac delta function described in detail in \cite{ch2001,vch2002}.

\subsection{Segmentation models incorporating noise distributions}

Studies \cite{pd2002,mrgg2004,s2013} investigated noisy image segmentation problems by using specific parameter estimation based on different noise distributions. All the related parameters are calculated via the maximum a posteriori probability (MAP) estimation from the viewpoint of Bayesian probability models. For example, estimation of variance information is used for images degraded with Gaussian noise; The square of image intensity value with capacity of enhancing weak properties is incorporated in the Rayleigh model; Models with great segmentation performance of dealing with Poisson and Gamma noises are built on the standard deviation and average. The general variational model is written as follows
\begin{eqnarray}\label{3}
E(\theta,\phi)=\alpha_1 \int_{\Omega} Q_1(x,\theta_1) \phi dx + \alpha_2 \int_{\Omega} Q_2(x,\theta_2)(1-\phi) dx + \gamma \int_{\Omega} |\nabla\phi| dx,
\end{eqnarray}
where $\alpha_1, \alpha_2, \gamma$ are positive penalty parameters, $\phi$ is a binary level set as defined in functional \reff{1}. Specific representations of function $Q$ derived from the maximum likelihood method and the computation of their related parameters are given in Table 1. $\theta =(\mu, \sigma)$ refers to the corresponding parameters of Function $Q$ need to be estimated.

\begin{table}[htbp]
\centering \tabcolsep 12pt
\begin{tabular}{ |c|c|c| }
\multicolumn{3}{c}{\textbf{TABLE 1} Potential functions of different noise distributions}\vspace{1mm}\\
\hline
\textbf{Functions} & \textbf{Gaussian noise} & \textbf{Rayleigh noise}  \\
\hline
$Q_{i(i=1,2)}$ & $\frac{1}{2}\log 2\pi+ \log \sigma_i+ \frac{(f-\mu_i)^2}{2\sigma_i^2}$  & $2 \log \sigma_i-\log f + \frac{f^2}{2\sigma_i^2}$  \\
\hline
\textbf{Parameters}  &	$\mu_i=\frac{\int_\Omega f \phi^{2-i}(1-\phi)^{i-1} dx}{\int_\Omega  \phi^{2-i}(1-\phi)^{i-1} dx}$  &	\multirow{2}{*}{$\sigma_i^2=\frac{\int_\Omega f^2 \phi^{2-i}(1-\phi)^{i-1} dx}{2\int_\Omega  \phi^{2-i}(1-\phi)^{i-1} dx}$}\\
$\theta_i=(\mu_i,\sigma_i)$ & $\sigma_i^2=\frac{\int_\Omega (f-\mu_i)^2 \phi^{2-i}(1-\phi)^{i-1} dx}{\int_\Omega  \phi^{2-i}(1-\phi)^{i-1} dx}$ & \\
\hline
\textbf{Functions} &	\textbf{Poisson noise} & \textbf{Gamma noise}  \\
\hline
 $Q_{i(i=1,2)}$ & $\sigma_i-f \log \sigma_i$	 &	$\frac{f}{\mu_i}+\log \mu_i$  \\
\hline
\textbf{Parameters}  &	\multirow{2}{*}{$\sigma_i=\frac{\int_\Omega f \phi^{2-i}(1-\phi)^{i-1} dx}{\int_\Omega  \phi^{2-i}(1-\phi)^{i-1} dx}$}  &	\multirow{2}{*}{$\mu_i=\frac{\int_\Omega f \phi^{2-i}(1-\phi)^{i-1} dx}{\int_\Omega  \phi^{2-i}(1-\phi)^{i-1} dx}$}\\
$\theta_i=(\mu_i,\sigma_i)$ & & \\
\hline
\end{tabular}
\end{table}

\subsection{Progressive hedging algorithm in stochastic programming}

As stated in \cite{rs2018}, $\Xi$ is a finite set with scenarios $\xi = (\xi_1, \xi_2, \ldots, \xi_N)$. Each scenario has a known probability $p(\xi) > 0$, and the sum of these probabilities is 1. Let $x(\cdot)$ be the mappings that designate responses to $\xi$ in the following form
\begin{eqnarray}\label{4}
x(\cdot):\xi\mapsto x(\xi)=(x_1(\xi), \ldots, x_N(\xi))\in \mathbb{R}^{n_1}\times\cdots \times \mathbb{R}^{n_N}=\mathbb{R}^{n}.
\end{eqnarray}
Here $\mathcal{L}$ is used to represent the linear space consisting of all such mappings $x(\cdot)$ from $\Xi$ to $\mathbb{R}^{n}$. $k=1, 2, \ldots$ denote iteration steps. Elements $x^{k+1}(\cdot)\in N$ and $w^{k+1}(\cdot)\in M$ are derived from elements $x^{k}(\cdot)\in N$ and $w^{k}(\cdot)\in M$. $x(\xi)$ denotes for each $\xi\in \Xi$ a variable in $\mathbb{R}^{n}$. $\hat{x}^{k}(\cdot)\in \mathcal{L}$ can be determined by solving a separate problem for each scenario $\xi$ to obtain $\hat{x}^{k}(\xi)$ as follows:
\begin{eqnarray}\label{5}
\hat{x}^{k}(\xi)=\arg\min_{x(\xi)\in C(\xi)}~\{g(x(\xi),\xi)+w^k(\xi)\cdot x(\xi)+\frac{r}{2}\|x(\xi)-x^k(\xi)\|^2\},\nonumber\\
\textrm{and~then}~x^{k+1}(\cdot)=P_N(\hat{x}^{k}(\cdot))~\textrm{and}~w^{k+1}(\cdot)=w^{k}(\cdot)+rP_M(\hat{x}^{k}(\cdot)).
\end{eqnarray}
$P_N$ and $P_M$ are the projection mappings onto the sub-spaces $N$ and $M$ of $\mathcal{L}$. The authors indicated that the vector $x(\xi)$ exists and can be uniquely determined with the reason that the proximal term guarantees the functional being minimized to be strongly convex. More details in terms of theorems and proofs can be found in \cite{rs2018}.

\section{Novel formulations for different segmentation purposes incorporating influence of stochastic noises via progressive hedging}

Motivated by the research using PHA to solve the minimization problem of stochastic programming, we aim to propose novel formulations tackling different segmentation issues in consideration of the advantages of Euler's elastica term and the influence of unknown noises. In this way, we not only can fulfill general segmentation tasks as well as the classic model, but also can deal with worse situations such as low quality images with large noises, absent boundaries, missing shapes or occlusion. Then we show how to implement PHA with developed ADMM-C algorithm to obtain the optimal solutions efficiently.

\subsection{Two-phase segmentation based application}

Based on two-phase Euler's elastica based segmentation formulated in \reff{1}, we propose a novel segmentation model incorporating influence of different noises expressed as the following stochastic programming (SP) form. The reconstructed contour can be obtained by minimizing the following energy functional with respect to $\phi(\xi)$.
\begin{eqnarray}\label{6}
&&\arg\min_{\theta_\xi,\phi_\xi\in\{0,1\}} \;\left\{E_{\scriptsize\textrm{SP-gray}}^{\scriptsize\textrm{Euler's~elastica}}(\theta_\xi,\phi_\xi)\right.\nonumber\\
&=&\alpha_1 \int_{\Omega} Q_1(x,\theta_1(\xi)) \phi(\xi) dx + \alpha_2 \int_{\Omega} Q_2(x,\theta_2(\xi))(1-\phi(\xi)) dx\\
&&\left.+ \int_{\Omega} (\alpha+\beta\left| \nabla \cdot \frac{\nabla\phi(\xi)}{|\nabla\phi(\xi)|}\right|) |\nabla\phi(\xi)| dx + \int_{\Omega} (v^k(\xi)\phi(\xi)+\frac{\tau}{2}(\phi(\xi)-\phi_\xi^k)^2)dx \right\} \nonumber
\end{eqnarray}
$\xi = (\xi_1, \xi_2, \cdots, \xi_N)$ represent different noise distributions and $Q_i(x, \theta_i(x))$ contain the stochastic information need to be estimated. $\phi(\xi)$ is the optimal solution of \reff{6} under distribution $\xi$. Here we continue the definition of $\phi(\xi)$ in \cite{ztch2013} using binary representation which can also be explained as a substitution $\phi=H(\varphi)$. As they described, this binary representation was originally used for finding the global minimizer. And it can also reduce the computational complexity to some extent such as avoiding the necessary calculation associated with level sets. The last two terms $v^k(\xi)\phi(\xi)+\frac{\tau}{2}(\phi(\xi)-\phi_\xi^k)^2$ added to sub-problem of $\phi(\xi)$ in functional \reff{6} can guarantee the mathematical convergence strictly.

According to the segmentation model for vector-valued images proposed in \cite{chs2000}, the averages of the data terms over all channels are used for coupling. Let $f =( f_1, f_2, \ldots, f_m)$ be a original color image defined on a domain $\Omega$. Then function $Q$ should be also in the multichannel form $(Q_1, Q_2, \ldots, Q_m)$. In fact, our proposed model used to solve color image segmentation can be stated as follows:
\begin{eqnarray}\label{7}
&&\arg\min_{\theta_\xi,\phi_\xi\in\{0,1\}} \;\left\{E_{\scriptsize\textrm{SP-color}}^{\scriptsize\textrm{Euler's~elastica}}(\theta_\xi,\phi_\xi)\right.\nonumber\\
&=&\alpha_1 \int_{\Omega} \sum_{l=1}^m Q_{1l}(x,\theta_{1l}(\xi)) \phi(\xi) dx + \alpha_2 \int_{\Omega} \sum_{l=1}^m Q_{2l}(x,\theta_{2l}(\xi))(1-\phi(\xi)) dx\\
&&\left.+ \int_{\Omega} (\alpha+\beta\left| \nabla \cdot \frac{\nabla\phi(\xi)}{|\nabla\phi(\xi)|}\right|) |\nabla\phi(\xi)| dx + \int_{\Omega} (v^k(\xi)\phi(\xi)+\frac{\tau}{2}(\phi(\xi)-\phi_\xi^k)^2)dx\right\} \nonumber
\end{eqnarray}
where $l =1, 2, \ldots, m$ denote the number of layers of a vector-valued image. In this way, we obtain novel Euler's elastica based formulations embedding stochastic noises for two-phase segmentation. In the following section we shall implement the calculation under PHA, which is one useful and effective tool for solving above multistage stochastic programming problem.

\subsection{PHA with developed ADMM-C algorithm for two-phase segmentation application}

In order to demonstrate the precise numerical procedure of PHA in solving \reff{6} and \reff{7}, we focus on the general model integrating the gray space and color space cases. Detailed proofs for convergence of this algorithm are provided in \cite{rs2018}. The original minimization problems \reff{6} and \reff{7} are based on separate sub optimization problems,
\begin{eqnarray}\label{8}
\arg\min_{\theta_\xi,\phi_\xi\in\{0,1\}} \;\{E_{\scriptsize\textrm{SP-general}}^{\scriptsize\textrm{Euler's~elastica}}(\theta_\xi,\phi_\xi)\} \Rightarrow\left\{\begin{array}{l} \arg\min_{\theta,\phi\in\{0,1\}} \;\{E_{\scriptsize\textrm{SP-general}}^{\scriptsize\textrm{Euler's~elastica}}(\theta_{\xi_1},\phi_{\xi_1})\} \vspace{1mm}\\ \argmin_{\theta,\phi\in\{0,1\}} \;\{E_{\scriptsize\textrm{SP-general}}^{\scriptsize\textrm{Euler's~elastica}}(\theta_{\xi_2},\phi_{\xi_2})\} \vspace{1mm}\\
\argmin_{\theta,\phi\in\{0,1\}}\;\{E_{\scriptsize\textrm{SP-general}}^{\scriptsize\textrm{Euler's~elastica}}(\theta_{\xi_3},\phi_{\xi_3})\} \vspace{1mm}\\
\argmin_{\theta,\phi\in\{0,1\}}\;\{E_{\scriptsize\textrm{SP-general}}^{\scriptsize\textrm{Euler's~elastica}}(\theta_{\xi_4},\phi_{\xi_4})\}
 \end{array}\right.
\end{eqnarray}
where $\xi = (\xi_1, \xi_2, \cdots, \xi_N)$ refer to noise distributions. For each distribution $\xi_i$ has a known probability $p(\xi_i)>0$ which can be set empirically through experiments, and the sum $\sum_i p(\xi_i)=1$ has to be guaranteed. According to PHA approach, $\phi_\xi$ can be obtained by the following steps.

\textbf{I}. First the minimization problems need to be solved separately, which are given in the right side of \reff{8}.
\begin{eqnarray}\label{9}
(\theta_{\xi_i}^{k+1},\phi_{\xi_i}^{k+1})&=&\arg\min_{\theta,\phi\in\{0,1\}} \;\left\{E_{\scriptsize\textrm{SP-general}}^{\scriptsize\textrm{Euler's~elastica}}(\theta_{\xi_i},\phi_{\xi_i})\right.\nonumber\\
&=&\alpha_1 \int_{\Omega} Q_{1}^*(x,\theta_{1}(\xi_i)) \phi(\xi_i) dx + \alpha_2 \int_{\Omega} Q_{2}^*(x,\theta_{2}(\xi_i))(1-\phi(\xi_i)) dx\\
&&\left.+ \int_{\Omega} (\alpha+\beta| \kappa(\xi_i)|) |\nabla\phi(\xi_i)| dx + \int_{\Omega} (v^k(\xi_i)\phi(\xi_i)+\frac{\tau}{2}(\phi(\xi_i)-\phi_\xi^k)^2)dx\right\} \nonumber
\end{eqnarray}
where $\kappa =\nabla \cdot \frac{\nabla\phi}{|\nabla\phi|}$. For the gray space issue, there is only one layer of the image information to be calculated, which means $Q^*(x, \theta(\xi_i))=Q(x, \theta(\xi_i))$. For the color space issue, $Q^*(x, \theta(\xi_i))$ should be substituted with the coupling terms $\sum_l Q_l(x, \theta_l(xi_i))$. $Q^*(x, \theta(\xi_i))$ is the potential function for specific noise distribution in each channel of the image. Table 1 shows the representations of $Q(x, \theta(\xi_i))$ and estimations of parameters $\theta(\xi_i)$. $\phi_{\xi_i}$ is the optimal solution of \reff{9} in distribution $\xi_i$.

\textbf{II}. Next all of the obtained optimal solutions are utilized to gain the final optimum $\phi_{\xi}$.
\begin{eqnarray}\label{10}
\phi_{\xi}^{k+1}=\sum_{i=1}^N p(\xi_i)\phi_{\xi_i}^{k+1},
\end{eqnarray}

\textbf{III}. At last, the sub problems' solutions $\phi^{k}({\xi_i})$ and the Lagrangian multipliers $v^k(\xi_i)$ need to be updated at the end of each iteration.
\begin{eqnarray}\label{11}
\left\{ \begin{array}{l} \phi^{k+1}({\xi_1})= \phi_{\xi}^{k+1}\vspace{1mm}\\ \phi^{k+1}({\xi_2})= \phi_{\xi}^{k+1}\vspace{1mm}\\ \vdots \\ \phi^{k+1}({\xi_N})= \phi_{\xi}^{k+1} \end{array} \right. ,\;\; \left\{ \begin{array}{l} v^{k+1}({\xi_1})= v^{k}({\xi_1})+\tau(\phi^{k+1}({\xi_1})- \phi_{\xi}^{k+1})\vspace{1mm}\\ v^{k+1}({\xi_2})= v^{k}({\xi_2})+\tau(\phi^{k+1}({\xi_2})- \phi_{\xi}^{k+1})\vspace{1mm}\\ \vdots \\ v^{k+1}({\xi_N})= v^{k}({\xi_N})+\tau(\phi^{k+1}({\xi_N})- \phi_{\xi}^{k+1}) \end{array} \right. .
\end{eqnarray}
Then updated $\phi^{k+1}({\xi_i})$, $v^{k+1}({\xi_i})$ and the parameters $\theta^{k+1}(\xi_i)$ derived from \reff{9} are passed to the next iteration from step \textbf{I}.

To solve the minimization problems \reff{9} separately, both simplification and effectiveness of the algorithm should be considered. In fact, there are three main computational difficulties in the functional as listed below, followed by the corresponding algorithm design in response.
\begin{itemize}
\item[(i)] One main challenge of the Euler's Elastica based functional is due to the non-smoothness and non-convexity of $g(\kappa)=\alpha+\beta |\kappa |$. As described in \cite{yk2016,bst2011,dgt2018}, it is more efficient when the curvature term $g(\kappa)$ is computed separately from the functional \reff{9}. Inspired by the concept of curvature weighted approach, we can rewrite functional \reff{9} as the following simplified version
    \begin{eqnarray}\label{12}
    (\theta_{\xi_i}^{k+1},\phi_{\xi_i}^{k+1})&=&\arg\min_{\theta,\phi\in\{0,1\}} \;\left\{E_{\scriptsize\textrm{SP-general}}^{\scriptsize\textrm{Euler's~elastica}}(\theta_{\xi_i},\phi_{\xi_i})\right.\\
    &=&\alpha_1 \int_{\Omega} Q_{1}^*(x,\theta_{1}(\xi_i)) \phi(\xi_i) dx + \alpha_2 \int_{\Omega} Q_{2}^*(x,\theta_{2}(\xi_i))(1-\phi(\xi_i)) dx\nonumber\\
    &&\left.+ \int_{\Omega} g(\kappa(\xi_i)) |\nabla\phi(\xi_i)| dx + \int_{\Omega} (v^k(\xi_i)\phi(\xi_i)+\frac{\tau}{2}(\phi(\xi_i)-\phi_\xi^k)^2)dx\right\} \nonumber
    \end{eqnarray}
    The proposed approach essentially reduces the minimization problem \reff{12} to a total variation type \cite{rof1992}. Here the case of division by zero in $g(\kappa)$ should be avoided. In practice, the denominator is often replaced by $|\nabla\phi|_\epsilon=\max(\epsilon,|\nabla\phi|)$ ($\epsilon$ is a arbitrarily small positive parameter). Then $g(\kappa(\xi_i))$ is represented as \\ $\nabla\cdot(\nabla\phi(\xi_i))/|\nabla\phi(\xi_i)|_\epsilon$.
\item[(ii)] Note that the binary constraint for $\phi$ will also cause non-convexity in sub problems \reff{12}. \cite{chen2006} demonstrated that certain non-convex minimization problems can be equivalent to the following convex minimization problems
    \begin{eqnarray}\label{13}
    (\theta_{\xi_i}^{k+1},\phi_{\xi_i}^{k+1})&=&\arg\min_{\theta,\phi\in[0,1]} \;\left\{E_{\scriptsize\textrm{SP-general}}^{\scriptsize\textrm{Euler's~elastica}}(\theta_{\xi_i},\phi_{\xi_i})\right.\\
    &=&\alpha_1 \int_{\Omega} Q_{1}^*(x,\theta_{1}(\xi_i)) \phi(\xi_i) dx + \alpha_2 \int_{\Omega} Q_{2}^*(x,\theta_{2}(\xi_i))(1-\phi(\xi_i)) dx\nonumber\\
    &&\left.+ \int_{\Omega} g(\kappa(\xi_i)) |\nabla\phi(\xi_i)| dx + \int_{\Omega} (v^k(\xi_i)\phi(\xi_i)+\frac{\tau}{2}(\phi(\xi_i)-\phi_\xi^k)^2)dx \right\}\nonumber
    \end{eqnarray}
    This convex minimization scheme could find global minimizers for \reff{12} by thresholding the solution of \reff{13}, which was classed as a continuous min-cut algorithm. Together with its equivalent form, the continuous max-flow algorithm, the min-cut algorithm has been proved to be an exact convex relaxation of the original problem as stated in \cite{ybt2010,b2014}.
\item[(iii)] Another critical issue for solving \reff{13} is the inevitable high order derivatives in numerical implementation, which is proven to be tedious and prone to errors. Then a developed ADMM-C algorithm is designed to each sub problem by introducing auxiliary variables, Lagrangian multipliers and an alternating directional optimization strategy.
\end{itemize}

Here the detailed implementation on solving each sub-problem by the ADMM-C algorithm will be presented. Firstly, some auxiliary variables are introduced, i.e., $\vec{w}(\xi_i)= [w_1(\xi_i), w_2(\xi_i)]^T$ with property $\vec{w}\approx \nabla\phi(\xi_i)$ and the Lagrangian multipliers  $\vec{\lambda}(\xi_i)= [\lambda_1(\xi_i), \lambda_2(\xi_i)]^T$. Based on this observation, we can transform \reff{13} into the following augmented Lagrangian functional
\begin{eqnarray}\label{14}
(\theta_{\xi_i}^{k+1},\phi_{\xi_i}^{k+1},\vec{w}_{\xi_i}^{k+1})&=&\arg\min_{\theta,\phi\in[0,1]} \;\left\{E_{\scriptsize\textrm{SP-general}}^{\scriptsize\textrm{Euler's~elastica}}(\theta_{\xi_i},\phi_{\xi_i})\right.\\
&=&\alpha_1 \int_{\Omega} Q_{1}^*(x,\theta_{1}(\xi_i)) \phi(\xi_i) dx + \alpha_2 \int_{\Omega} Q_{2}^*(x,\theta_{2}(\xi_i))(1-\phi(\xi_i)) dx\nonumber\\
&&+ \int_{\Omega} g(\kappa(\xi_i)) |\vec{w}(\xi_i)| dx + \int_{\Omega} (v^k(\xi_i)\phi(\xi_i)+\frac{\tau}{2}(\phi(\xi_i)-\phi_\xi^k)^2)dx \nonumber\\
&&\left.+ \int_{\Omega} (\vec{\lambda}(\xi_i)\cdot(\vec{w}(\xi_i)- \nabla\phi(\xi_i))+\frac{\mu}{2}(\vec{w}(\xi_i)- \nabla\phi(\xi_i))^2)dx\right\}\nonumber
\end{eqnarray}
where $\mu$ is a positive penalty parameter. It is worth noting that this kind of simple structure of \reff{14} requires fewer variables and parameters compared with other works [11,18,19] using ADMM to deal with the curvature term directly. After the initialization of $\phi^0(\xi_i)$, $\vec{w}^0(\xi_i)$ and $\vec{\lambda}^0(\xi_i)$, a minimization problem is carried out in each step with respect to one variable while keeping other variables fixed temporarily. When alternative optimization for all the variables is completed, the Lagrangian multipliers will be updated subsequently. This gives
\begin{eqnarray}\label{15}
\theta_{\xi_i}^{k+1}=\arg\min_{\theta} \left\{ \alpha_1 \int_{\Omega} Q_{1}^*(x,\theta_{1}(\xi_i)) \phi^k(\xi_i) dx + \alpha_2 \int_{\Omega} Q_{2}^*(x,\theta_{2}(\xi_i))(1-\phi^k(\xi_i)) dx \right\},
\end{eqnarray}
\begin{eqnarray}\label{16}
\phi_{\xi_i}^{k+1}&=&\arg\min_{\phi\in[0,1]} \left\{\alpha_1 \int_{\Omega} Q_{1}^*(x,\theta_{1}^{k+1}(\xi_i)) \phi(\xi_i) dx + \alpha_2 \int_{\Omega} Q_{2}^*(x,\theta_{2}^{k+1}(\xi_i))(1-\phi(\xi_i)) dx\right.\nonumber\\
&&+ \int_{\Omega} (v^k(\xi_i)\phi(\xi_i)+\frac{\tau}{2}(\phi(\xi_i)-\phi_\xi^k)^2)dx \\
&&\left.+ \int_{\Omega} (\vec{\lambda}^k(\xi_i)\cdot(\vec{w}^k(\xi_i)- \nabla\phi(\xi_i))+\frac{\mu}{2}(\vec{w}^k(\xi_i)- \nabla\phi(\xi_i))^2)dx\right\}\nonumber
\end{eqnarray}
\begin{eqnarray}\label{17}
\vec{w}_{\xi_i}^{k+1}&=&\argmin_{\vec{w}} \left\{\int_{\Omega} g(\kappa^{k+1}(\xi_i)) |\vec{w}(\xi_i)| dx +\int_{\Omega} \vec{\lambda}^k(\xi_i)\cdot(\vec{w}(\xi_i)- \nabla\phi^{k+1}(\xi_i))\right.\\ &&\left.+\frac{\mu}{2}(\vec{w}(\xi_i)- \nabla\phi^{k+1}(\xi_i))^2dx\right\},\quad \textrm{where}~g(\kappa^{k+1}(\xi_i))=\nabla\cdot\frac{\nabla\phi^{k+1}_{\xi_i}}{|\nabla\phi^{k+1}_{\xi_i}|_\epsilon},\nonumber
\end{eqnarray}
\begin{eqnarray}\label{18}
\vec{\lambda}_{\xi_i}^{k+1}=\vec{\lambda}_{\xi_i}^{k}+\mu(\vec{w}^{k+1}({\xi_i})-\nabla\phi^{k+1}(\xi_i)).
\end{eqnarray}
\textbf{To obtain} $\theta^{k+1}=(\mu, \sigma)$: In the $k+1$ step of the proposed ADMM-C, the average image intensity values $\mu_{\xi_i}$ as well as variances $\sigma_{\xi_i}$ in the foreground and background can be obtained by using the standard variational method for \reff{15}. Table 1 gives all the solutions for distributions $\xi = (\xi_1, \xi_2, \ldots, \xi_N)$.\\
\textbf{To obtain} $\phi^{k+1}$: Optimal value of $\phi_{\xi_i}$ for a certain distribution $\xi_i$ is obtained by solving the minimization of \reff{16} with respect to $\phi(\xi_i)$. We can get the update rule based on the corresponding Euler-Lagrange equations
\begin{eqnarray}\label{19}
(-\mu \Delta+v^k(\xi_i)+\tau)\phi(\xi_i)=\tau\phi_\xi^k-r_{\xi_i}(\theta_1^{k+1},\theta_2^{k+1})-\nabla\cdot \vec{\lambda}^{k}({\xi_i})-\mu\nabla\cdot \vec{w}^{k}({\xi_i}),
\end{eqnarray}
where $r_{\xi_i}(\theta_1^{k+1},\theta_2^{k+1})=\alpha_1 Q_{1}^*(x,\theta_{1}^{k+1}(\xi_i)) - \alpha_2 Q_{2}^*(x,\theta_{2}^{k+1}(\xi_i))$. Like in \cite{mps2014}, equation \reff{19} is a screened Poisson equation for which Fast Fourier transform (FFT) \cite{dgt2018,mps2014} is a well-known solver with very low computational cost for imaging problems. Here FFT is applied for further improving the calculation efficiency. Equation \reff{19} can be rewritten as
\begin{eqnarray}\label{20}
F^*LF\phi(\xi_i)=\tau\phi_\xi^k-r_{\xi_i}(\theta_1^{k+1},\theta_2^{k+1})-\nabla\cdot \vec{\lambda}^{k}({\xi_i})-\mu\nabla\cdot \vec{w}^{k}({\xi_i}),
\end{eqnarray}
where $L=-\mu F\Delta F^*+v^k(\xi_i)+\tau$ and $F^*$ is the discrete inverse Fourier transform. Then we can obtain $\phi_{\xi_i}^{k+1}$ as follows
\begin{eqnarray}\label{21}
\phi_{\xi_i}^{k+1}=F^*(L^{-1}F(\tau\phi_\xi^k-r_{\xi_i}(\theta_1^{k+1},\theta_2^{k+1})-\nabla\cdot \vec{\lambda}^{k}({\xi_i})-\mu\nabla\cdot \vec{w}^{k}({\xi_i}))).
\end{eqnarray}
\textbf{To obtain} $\vec{w}^{k+1}$: The minimization problem \reff{17} of $\vec{w}$ can be solved via the generalized soft thresholding formula \cite{tll2018,tlp2018}, which is given by
\begin{eqnarray}\label{22}
\vec{w}^{k+1}_{\xi_i}=\max\left( |\nabla\phi_{\xi_i}^{k+1}-\frac{\vec{\lambda}^{k}({\xi_i})}{\mu}|-\frac{g(\kappa^{k+1}(\xi_i))}{\mu} ,0\right)\frac{\nabla\phi_{\xi_i}^{k+1}-\frac{\vec{\lambda}^{k}({\xi_i})}{\mu}}{|\nabla\phi_{\xi_i}^{k+1}-\frac{\vec{\lambda}^{k}({\xi_i})}{\mu}|}.
\end{eqnarray}

For clarity, we present the overall algorithm for the two-phase Euler's elastica based segmentation in stochastic programming in a pseudo code format as follows.
\begin{table}[H]
\centering  \tabcolsep 10pt
\begin{tabular}{l}
\hline
\textbf{\small Algorithm 1. Computing framework for \reff{6} and \reff{7} via PHA }  \\
\hline
\textbf{Input:} $\phi^0(\xi_i), p(\xi_i),v^0(\xi_i), \alpha, \beta, \tau, \alpha_1, \alpha_2$\vspace{2mm} \\
$\quad$ \textbf{for} $k\geq 1$, do the following steps recurrently\vspace{2mm} \\
$\quad\quad$ 1: Obtain $\phi^{k+1}_{\xi_i}$ via \textbf{Algorithm 2} \vspace{2mm}\\
$\quad\quad$ 2: Update $\phi^{k+1}_{\xi}$ via Equation \reff{10} \vspace{2mm}\\
$\quad\quad$ 3: Update $\phi^{k+1}({\xi_i})$, $v^{k+1}(\xi_i)$ via Equation \reff{11} \vspace{2mm}\\
$\quad\quad$ 4: \textbf{if} \emph{some stopping criteria (given in Section 3.5) are satisfied} \textbf{break} \vspace{2mm}\\
\textbf{Return} optimal value $\phi^{k+1}_{\xi}$ after thresholding \\
\hline
\end{tabular}
\end{table}

\begin{table}[H]
\centering  \tabcolsep 8pt
\begin{tabular}{l}
\hline
\textbf{\small Algorithm 2 Detailed implementation for step 1 in Algorithm 1 via ADMM-C}  \\
\hline
\textbf{If} $k=1$\vspace{2mm} \\
$\quad$ input $\vec{w}^0(\xi_i), \vec{\lambda}^0(\xi_i), \mu$ \vspace{2mm} \\
\textbf{else} solve the following problems alternatively \vspace{2mm} \\
$\quad$ 1: Update $\theta^{k+1}_{\xi_i}$ according to distribution laws  \vspace{2mm}\\
$\quad$ 2: Update $\phi^{k+1}_{\xi}$ via minimization problem \reff{16}  \vspace{2mm}\\
$\quad$ 3: Update $\vec{w}^{k+1}_{\xi_i}$ via minimization problem \reff{17} \vspace{2mm}\\
$\quad$ 4: Update $\vec{\lambda}^{k+1}_{\xi_i}$ via \reff{18} using gradient ascent method \\
\hline
\end{tabular}
\end{table}

The idea of the ADMM-C method is intentionally applied for the main challenge of the non-convex, non-smooth and non-linear problems in Euler's elastics and it has attracted extensive attention. ADMM algorithm has been given analytical properties in \cite{hl2017,lsg2017} and many other applications \cite{mgkr2015,m2015} are provided for which many similar algorithms are developed and successfully used to achieve excellent performances via solving a variety of non-convex problems. In this paper, we adopt a similar idea as in \cite{yk2016,dgt2018} to design a new algorithm to deal with the sub-problems derived from PHA framework.

\subsection{Segmentation with depth based application}

In this part, we intend to use similar stochastic programming skills as the ones applied in the two-phase issue. Based on original segmentation with depth model for gray space reviewed in Section 2.1, we can establish the energy functional by introducing random noise set as follows
{\small \begin{eqnarray}\label{23}
&&\arg\min_{\theta_\xi,\phi_\xi\in\{0,1\}} \;\left\{E_{\scriptsize\textrm{SP-gray}}^{\scriptsize\textrm{Depth}}(\theta_{\xi},\phi_{\xi})\right.\\
&=&\sum_{h=1}^n \int_{\Omega} (\alpha+\beta|\nabla\cdot \frac{\nabla\phi_h(\xi)}{|\nabla\phi_h(\xi)|} |)|\nabla\phi_h(\xi)|dx+\sum_{h=1}^n \int_{\Omega} Q_h(x,\theta_h(\xi))\phi_h(\xi)\prod\limits_{j=1}^{h-1} (1-\phi_j(\xi))dx \nonumber\\
&&\left.+ \int_{\Omega} Q_{n+1}(x,\theta_{n+1}(\xi))\prod\limits_{j=1}^{n} (1-\phi_j(\xi))dx+\sum_{h=1}^n \int_{\Omega} (v_h^k(\xi)\cdot\phi_h(\xi)+\frac{\tau}{2}(\phi_h(\xi)-\phi_h^k(\xi))^2)dx\right\}\nonumber
\end{eqnarray}}
where $h=1, 2, \ldots, n$ denote the number of objects in the image, and $\phi_{n+1}(\xi)=1$ is set only for consistency of description. We can still extend above segmentation with depth incorporating stochastic noises model to multichannel case. Analog to the coupling approach used in \reff{7}, the formulation is written as
\begin{eqnarray}\label{24}
&&\arg\min_{\theta_\xi,\phi_\xi\in\{0,1\}} \;\left\{E_{\scriptsize\textrm{SP-color}}^{\scriptsize\textrm{Depth}}(\theta_{\xi},\phi_{\xi})\right.\\
&=&\sum_{h=1}^n \int_{\Omega} (\alpha+\beta|\nabla\cdot \frac{\nabla\phi_h(\xi)}{|\nabla\phi_h(\xi)|} |)|\nabla\phi_h(\xi)|dx\nonumber\\
&&+\sum_{h=1}^n \int_{\Omega} \sum_{l=1}^m Q_{hl}(x,\theta_{hl}(\xi))\phi_h(\xi)\prod\limits_{j=1}^{h-1} (1-\phi_j(\xi))dx \nonumber\\
&&+ \int_{\Omega} \sum_{l=1}^m Q_{(n+1)l}(x,\theta_{(n+1)l}(\xi))\prod\limits_{j=1}^{n} (1-\phi_j(\xi))dx\nonumber\\
&&\left.+\sum_{h=1}^n \int_{\Omega} (v_h^k(\xi)\cdot\phi_h(\xi)+\frac{\tau}{2}(\phi_h(\xi)-\phi_{h(\xi)}^k)^2)dx\right\}\nonumber
\end{eqnarray}
In order to describe the calculation procedure explicitly, we plan to conduct on the general model integrating the gray space and color space cases. According to the curvature-weighted approach used in \reff{12}, the simplified version can be directly written as
\begin{eqnarray}\label{25}
&&\arg\min_{\theta_\xi,\phi_\xi\in\{0,1\}} \;\left\{E_{\scriptsize\textrm{SP-general}}^{\scriptsize\textrm{Depth}}(\theta_{\xi},\phi_{\xi})\right.\\
&=&\sum_{h=1}^n \int_{\Omega} g(\kappa_h(\xi))|\nabla\phi_h(\xi)|dx+\sum_{h=1}^n \int_{\Omega} Q_h^*(x,\theta_h(\xi))\chi_h(\xi)dx \nonumber\\
&&\left.+ \int_{\Omega} Q_{n+1}^*(x,\theta_{n+1}(\xi))\chi_{n+1}(\xi)dx+\sum_{h=1}^n \int_{\Omega} (v_h^k(\xi)\cdot\phi_h(\xi)+\frac{\tau}{2}(\phi_h(\xi)-\phi_{h(\xi)}^k)^2)dx\right\}\nonumber
\end{eqnarray}
where $g(\kappa_h(\xi))=\nabla\cdot (\nabla\phi_h(\xi))/|\nabla\phi_h(\xi)|_\epsilon)$ and the definition of $|\nabla\phi_h(\xi)|_\epsilon$ is given in (). $Q_h^*(x, \theta_h(\xi))$ is $Q_h(x, \theta_h(\xi))$ for gray space issue and $\sum_l Q_{hl}(x, \theta_{hl}(\xi))$ for color space issue. And the characteristic function for the $h$-th region reads $\chi_h(\xi)=\phi_h(x)\prod_{j=1:h-1}(1-\phi_j(\xi))$. Particularly, $\chi_{n+1}(\xi)=\phi_{n+1}(\xi)\prod_{j=1:n}(1-\phi_j(\xi))=\prod_{j=1:n}(1-\phi_j(\xi))$ representing the $(n+1)$-th region: background. Next section we shall calculate above general formulation under PHA with detailed implementation.

\subsection{PHA with developed ADMM-C algorithm for segmentation with depth application}

For the numerical part, we shall try to use similar ideas as were used in the sophisticated two-phase imaging tasks. The main program loop for PHA is shown below

\textbf{I}. First we need to solve the sub minimization problems of (25) separately, which gives
\begin{eqnarray}\label{26}
&&(\theta_{h(\xi_i)}^{k+1},\phi_{h(\xi_i)}^{k+1})\nonumber\\
&=&\arg\min_{\theta_h,\phi_h\in\{0,1\}} \;\left\{E_{\scriptsize\textrm{SP-general}}^{\scriptsize\textrm{Depth}}(\theta_{h(\xi_i)},\phi_{h(\xi_i)})\right.\\
&=&\sum_{h=1}^n \int_{\Omega} g(\kappa_h(\xi_i))|\nabla\phi_h(\xi_i)|dx+\sum_{h=1}^n \int_{\Omega} Q_h^*(x,\theta_h(\xi_i))\chi_h(\xi_i)dx \nonumber\\
&&\left.+ \int_{\Omega} Q_{n+1}^*(x,\theta_{n+1}(\xi_i))\chi_{n+1}(\xi_i)dx+\sum_{h=1}^n \int_{\Omega} (v_h^k(\xi_i)\cdot\phi_h(\xi_i)+\frac{\tau}{2}(\phi_h(\xi_i)-\phi_{h(\xi)}^k)^2)dx\right\}\nonumber
\end{eqnarray}

\textbf{II}. Next all of the obtained optimal solutions $(\phi_{h(\xi_1)}, \phi_{h(\xi_2)}, ..., \phi_{h(\xi_N)})$ are utilized to gain the final optimum $\phi_{h(\xi)}$.
\begin{eqnarray}\label{27}
\phi_{h(\xi)}^{k+1} = \sum_{i=1}^N p(\xi_i)\phi_{h(\xi_i)}^{k+1}.
\end{eqnarray}

\textbf{III}. At last, the sub problems' solutions $\phi_h^k(\xi_i)$ and the Lagrangian multipliers $v_h^k(\xi_i)$ need to be updated at the end of each iteration.
\begin{eqnarray}\label{28}
\left\{ \begin{array}{l} \phi_h^{k+1}({\xi_1})= \phi_{h(\xi)}^{k+1}\vspace{1mm}\\ \phi^{k+1}_h({\xi_2})= \phi_{h(\xi)}^{k+1}\vspace{1mm}\\ \vdots \\ \phi_h^{k+1}({\xi_N})= \phi_{h(\xi)}^{k+1} \end{array} \right. ,\;\; \left\{ \begin{array}{l} v_h^{k+1}({\xi_1})= v_h^{k}({\xi_1})+\tau(\phi_h^{k+1}({\xi_1})- \phi_{h(\xi)}^{k+1})\vspace{1mm}\\ v_h^{k+1}({\xi_2})= v_h^{k}({\xi_2})+\tau(\phi_h^{k+1}({\xi_2})- \phi_{h(\xi)}^{k+1})\vspace{1mm}\\ \vdots \\ v_h^{k+1}({\xi_N})= v_h^{k}({\xi_N})+\tau(\phi_h^{k+1}({\xi_N})- \phi_{h(\xi)}^{k+1}) \end{array} \right. .
\end{eqnarray}
Then updated $\phi_h^{k+1}({\xi_i})$, $v_h^{k+1}({\xi_i})$ and the parameters $\theta^{k+1}(\xi_i)$ derived from \reff{26} are passed to the next iteration from step \textbf{I}.

In order to solve the minimization problems \reff{26} in step \textbf{I} separately, we will show how to apply the designed ADMM-C algorithm. Note that there are n binary functions $(\phi_1(\xi_i), \phi_2(\xi_i), \ldots, \phi_n(\xi_i))$ need to be obtained for one specific noise distribution. Thus $n$ auxiliary variables $(\vec{w}_1(\xi_i), \vec{w}_2(\xi_i), \ldots, \vec{w}_n(\xi_i))$ are introduced and each component is defined as $\vec{w}_h(\xi_i)= [\vec{w}_{h1}(\xi_i), \vec{w}_{h2}(\xi_i)]^T$ with property $\vec{w}_{h}\approx \nabla\phi_h(\xi_i)$. Likewise, $n$ Lagrangian multipliers $(\vec{\lambda}_1(\xi_i), \vec{\lambda}_2(\xi_i), \ldots, \vec{\lambda}_n(\xi_i))$ are also brought in with  $\vec{\lambda}_h(\xi_i)= [\vec{\lambda}_{h1}(\xi_i), \vec{\lambda}_{h2}(\xi_i)]^T$. Based on the convex relaxation method, \reff{26} is rewritten into the following augmented Lagrangian functional
\begin{eqnarray}\label{29}
&&(\theta_{h(\xi_i)}^{k+1},\phi_{h(\xi_i)}^{k+1},\vec{w}_{h(\xi_i)}^{k+1})\nonumber\\
&=&\arg\min_{\theta_h,\phi_h\in\{0,1\}} \;\left\{E_{\scriptsize\textrm{SP-general}}^{\scriptsize\textrm{Depth}}(\theta_{h(\xi_i)},\phi_{h(\xi_i)})\right.\\
&=&\sum_{h=1}^n \int_{\Omega} g(\kappa_h(\xi_i))|\nabla\phi_h(\xi_i)|dx+\sum_{h=1}^n \int_{\Omega} Q_h^*(x,\theta_h(\xi_i))\chi_h(\xi_i)dx \nonumber\\
&&+ \int_{\Omega} Q_{n+1}^*(x,\theta_{n+1}(\xi_i))\chi_{n+1}(\xi_i)dx+\sum_{h=1}^n \int_{\Omega} v_h^k(\xi_i)\cdot\phi_h(\xi_i)+\frac{\tau}{2}(\phi_h(\xi_i)-\phi_{h(\xi)}^k)^2dx\nonumber\\
&&\left.+\sum_{h=1}^n \int_{\Omega} \vec{\lambda}_h(\xi_i)\cdot(\vec{w}_{h}(\xi_i)- \nabla\phi_h(\xi_i))+\frac{\mu}{2} (\vec{w}_{h}(\xi_i)- \nabla\phi_h(\xi_i))^2dx \right\}\nonumber
\end{eqnarray}
where $\mu$ is a positive penalty parameter. Here $h=1, 2, \ldots, n$ refer to the number of binary level set functions and $i=1, 2, \ldots, N$ refer to the number of noise distributions. In order to solve \reff{29} efficiently with the ADMM-C, we first initialize the unknown $\phi_h^0(\xi_i)$, $\vec{w}_{h}^0(\xi_i)$ and $\vec{\lambda}_{h}^0(\xi_i)$ at the initial iterative step $k=0$, then, we solve some minimization problems with espect to only one kind of unknowns while other ones are temporarily fixed at each step from $k$-th to $(k+1)$-th until convergence is reached. With this alternating direction optimization strategy, we can divide the optimization problem \reff{29} into three minimization problems in the iterative process from $k$-th to $(k+1)$-th step:
\begin{eqnarray}\label{30}
\theta_{h(\xi_i)}^{k+1}=\arg\min_{\theta_h} \left\{ \sum_{h=1}^n \int_{\Omega} Q_{h}^*(x,\theta_{h}(\xi_i)) \chi_h^k(\xi_i) dx + \int_{\Omega} Q_{n+1}^*(x,\theta_{n+1}(\xi_i)) \chi^k_{n+1}(\xi_i) dx \right\},
\end{eqnarray}
\begin{eqnarray}\label{31}
\phi_{h(\xi_i)}^{k+1}&=&\arg\min_{\phi\in[0,1]} \left\{\sum_{h=1}^n \int_{\Omega} Q_{h}^*(x,\theta_{h}^{k+1}(\xi_i)) \chi_h(\xi_i) dx + \int_{\Omega} Q_{n+1}^*(x,\theta_{n+1}^{k+1}(\xi_i))\chi_{n+1}(\xi_i) dx\right.\nonumber\\
&&+ \sum_{h=1}^n\int_{\Omega} (v_h^k(\xi_i)\phi_h(\xi_i)+\frac{\tau}{2}(\phi_h(\xi_i)-\phi_{h(\xi)}^k)^2)dx \\
&&\left.+ \sum_{h=1}^n \int_{\Omega} (\vec{\lambda}_h^k(\xi_i)\cdot(\vec{w}_h^k(\xi_i)- \nabla\phi_h(\xi_i))+\frac{\mu}{2}(\vec{w}_h^k(\xi_i)- \nabla\phi_h(\xi_i))^2)dx\right\}\nonumber
\end{eqnarray}
\begin{eqnarray}\label{32}
\vec{w}_{h(\xi_i)}^{k+1}&=&\arg\min_{\vec{w}_h} \left\{\sum_{h=1}^n\int_{\Omega} g(\kappa_h^{k+1}(\xi_i)) |\vec{w}_h(\xi_i)| dx +\sum_{h=1}^n\int_{\Omega} \vec{\lambda}_h^k(\xi_i)\cdot(\vec{w}_h(\xi_i)- \nabla\phi_h^{k+1}(\xi_i))\right.\nonumber\\ &&\left.+\frac{\mu}{2}(\vec{w}_h(\xi_i)- \nabla\phi_h^{k+1}(\xi_i))^2dx\right\},\quad \textrm{where}~g(\kappa_h^{k+1}(\xi_i))=\nabla\cdot\frac{\nabla\phi^{k+1}_{h(\xi_i)}}{|\nabla\phi^{k+1}_{h(\xi_i)}|_\epsilon},
\end{eqnarray}
\begin{eqnarray}\label{33}
\vec{\lambda}_{h(\xi_i)}^{k+1}=\vec{\lambda}_{h(\xi_i)}^{k}+\mu(\vec{w}_h^{k+1}({\xi_i})-\nabla\phi_h^{k+1}(\xi_i)).
\end{eqnarray}
\textbf{To obtain} $\theta_h^{k+1}=(\mu_h, \sigma_h)$: The average image intensity values $\mu_h^{k+1}(\xi_i)$ as well as variances $\sigma_h^{k+1}(\xi_i)$ in the foreground and background can be obtained by using the standard variational method based on \reff{30}, which are given by the following equations

\begin{table}[H]
\centering \tabcolsep 12pt
\begin{tabular}{ |c|c|c| }
\multicolumn{3}{c}{\textbf{TABLE 2} Potential functions of different noise distributions for}\vspace{1mm}\\
\multicolumn{3}{c}{segmentation with depth application}\vspace{1mm}\\
\hline
\textbf{Functions} & \textbf{Gaussian noise} & \textbf{Rayleigh noise}  \\
\hline
$Q_{h'}$ & \multirow{2}{*}{$\frac{1}{2}\log 2\pi+ \log \sigma_{h'}+ \frac{(f-\mu_{h'})^2}{2\sigma_{h'}^2}$}  & \multirow{2}{*}{$2 \log \sigma_{h'}-\log f + \frac{f^2}{2\sigma_{h'}^2}$}  \\
$(h'=1,\ldots,n+1)$ & &\\
\hline
\textbf{Parameters}  &	$\mu_{h'}=\frac{\int_\Omega f \chi_{h'} dx}{\int_\Omega  \chi_{h'} dx}$  &	\multirow{2}{*}{$\sigma_{h'}^2=\frac{\int_\Omega f^2 \chi_{h'} dx}{2\int_\Omega  \chi_{h'} dx}$}\\
$\theta_{h'}=(\mu_{h'},\sigma_{h'})$ & $\sigma_{h'}^2=\frac{\int_\Omega (f-\mu_{h'})^2 \chi_{h'} dx}{\int_\Omega \chi_{h'} dx}$ & \\
\hline
\textbf{Functions} &	\textbf{Poisson noise} & \textbf{Gamma noise}  \\
\hline
 $Q_{h'}$ & \multirow{2}{*}{$\sigma_{h'}-f \log \sigma_{h'}$}	 &	\multirow{2}{*}{$\frac{f}{\mu_{h'}}+\log \mu_{h'}$}  \\
$(h'=1,\ldots,n+1)$ & &\\
\hline
\textbf{Parameters}  &	\multirow{2}{*}{$\sigma_{h'}=\frac{\int_\Omega f \chi_{h'} dx}{\int_\Omega  \chi_{h'} dx}$}  &	\multirow{2}{*}{$\mu_{h'}=\frac{\int_\Omega f \chi_{h'} dx}{\int_\Omega  \chi_{h'} dx}$}\\
$\theta_{h'}=(\mu_{h'},\sigma_{h'})$ & & \\
\hline
\end{tabular}
\end{table}
\textbf{To obtain} $\phi_h^{k+1}$: For the minimization problem \reff{31} with respect to the function $\phi_{h(\xi_i)}$, the corresponding Euler-Lagrange equation is given as
\begin{eqnarray}\label{34}
&&(-\mu \Delta+v_h^k(\xi_i)+\tau)\phi_h(\xi_i)\nonumber\\
&=&-Q^*_h(x,\theta_h^{k+1}(\xi_i))\prod_{j=1}^{h-1}(1-\phi_j(\xi_i))+\tau\phi_{h(\xi)}^k-\nabla\cdot \vec{\lambda}_h^{k}({\xi_i})-\mu\nabla\cdot \vec{w}_h^{k}({\xi_i})\nonumber\\
&&+\sum_{s=h+1}^{n+1}\{Q_s^*(x,\theta_s^{k+1}(\xi_i))\phi_s(\xi_i)\prod_{j=1}^{h-1}(1-\phi_j(\xi_i))\prod_{j=h+1}^{s-1}(1-\phi_j(\xi_i))\}
\end{eqnarray}
Though above equation is more complicated, FFT can be still applied here for accelerating the calculation. Equation \reff{34} can be rewritten as
\begin{eqnarray}\label{35}
F^*LF\phi_h(\xi_i)=\tau\phi_{h(\xi)}^k-\Lambda_h(\xi_i)-\nabla\cdot \vec{\lambda}_h^{k}({\xi_i})-\mu\nabla\cdot \vec{w}_h^{k}({\xi_i}),
\end{eqnarray}
where $L=-\mu F\Delta F^*+v^k(\xi_i)+\tau$ and $F^*$ is the discrete inverse Fourier transform and
\begin{eqnarray*}
\Lambda_h(\xi_i)&=&Q^*_h(x,\theta_h^{k+1}(\xi_i))\prod_{j=1}^{h-1}(1-\phi_j(\xi_i))\\
&&-\sum_{s=h+1}^{n+1}\{Q_s^*(x,\theta_s^{k+1}(\xi_i))\phi_s(\xi_i)\prod_{j=1}^{h-1}(1-\phi_j(\xi_i))\prod_{j=h+1}^{s-1}(1-\phi_j(\xi_i)).
\end{eqnarray*}
Then we can obtain optimal value  of $\phi_{h(\xi_i)}$ as follows
\begin{eqnarray}\label{36}
\phi_{h(\xi_i)}^{k+1}=F^*(L^{-1}F(\tau\phi_{h(\xi)}^k-\Lambda_h(\xi_i)-\nabla\cdot \vec{\lambda}_h^{k}({\xi_i})-\mu\nabla\cdot \vec{w}_h^{k}({\xi_i}))).
\end{eqnarray}
\textbf{To obtain} $\vec{w}_h^{k+1}$: The calculation result of $\vec{w}$ minimization problem \reff{32} can be obtained via the generalized soft thresholding formula as
\begin{eqnarray}\label{37}
\vec{w}^{k+1}_{h(\xi_i)}=\max\left( |\nabla\phi_{h(\xi_i)}^{k+1}-\frac{\vec{\lambda}_h^{k}(\xi_i)}{\mu}|-\frac{g(\kappa_h^{k+1}(\xi_i))}{\mu} ,0\right)\frac{\nabla\phi_{h(\xi_i)}^{k+1}-\frac{\vec{\lambda}_h^{k}({\xi_i})}{\mu}}{|\nabla\phi_{h(\xi_i)}^{k+1}-\frac{\vec{\lambda}_h^{k}({\xi_i})}{\mu}|}.
\end{eqnarray}

For clarity, the overall algorithm for the Euler's elastica based segmentation with depth in stochastic programming in a pseudo code format is presented as follows.
\begin{table}[H]
\centering  \tabcolsep 10pt
\begin{tabular}{l}
\hline
\textbf{\small Algorithm 3 Computing framework for \reff{23} and \reff{24} via PHA }  \\
\hline
\textbf{Input:} $\phi_h^0(\xi_i)$ $(h=1,\ldots,n)$, $p(\xi_i),v_h^0(\xi_i), \alpha, \beta, \tau$\vspace{2mm} \\
$\quad$ \textbf{for} $k\geq 1$, do the following steps in turn\vspace{2mm} \\
$\quad\quad$ 1: Obtain $\phi^{k+1}_{h(\xi_i)}$ via \textbf{Algorithm 4} \vspace{2mm}\\
$\quad\quad$ 2: Update $\phi^{k+1}_{h(\xi)}$ via Equation \reff{27} \vspace{2mm}\\
$\quad\quad$ 3: Update $\phi_h^{k+1}({\xi_i})$, $v_h^{k+1}(\xi_i)$ via Equation \reff{28} \vspace{2mm}\\
$\quad\quad$ 4: \textbf{if} \emph{some stopping criteria (given in Section 3.5) are satisfied} \textbf{break} \vspace{2mm}\\
\textbf{Return} optimal value $(\phi_{1(\xi)}^{k+1},\phi_{2(\xi)}^{k+1},\ldots,\phi_{n(\xi)}^{k+1})$ after thresholding \\
\hline
\end{tabular}
\end{table}

\begin{table}[H]
\centering  \tabcolsep 8pt
\begin{tabular}{l}
\hline
\textbf{\small Algorithm 4 Detailed implementation for step 2 in Algorithm 3 via ADMM-C}  \\
\hline
\textbf{If} $k=1$\vspace{1mm} \\
$\quad$ input $\vec{w}_h^0(\xi_i)$, and $\vec{\lambda}_h^0(\xi_i) (h=1,\ldots,n), \mu$ \vspace{2mm}\\
\textbf{else} solve the following problems alternatively \vspace{2mm} \\
$\quad$ 1: Update $\theta^{k+1}_{h(\xi_i)}$ according to distribution laws  \vspace{2mm}\\
$\quad$ 2: Update $\phi^{k+1}_{h(\xi_i)}$ via minimization problem \reff{31}  \vspace{2mm}\\
$\quad$ 3: Update $\vec{w}^{k+1}_{h(\xi_i)}$ via minimization problem \reff{32} \vspace{2mm}\\
$\quad$ 4: Update $\vec{\lambda}^{k+1}_{h(\xi_i)}$ via \reff{33} using gradient ascent method \\
\hline
\end{tabular}
\end{table}

\subsection{Termination criteria}

The stopping criteria for our entire algorithm are described in this section. As described in \cite{tlp2018,btz2017}, the iterations need to be terminated when the following criteria are satisfied:
\begin{itemize}
\item \textbf{For major framework PHA in Algorithm 1 and Algorithm 3:} During iteration, the constraint errors of $(\phi_{\xi_i}-\phi_{\xi})$, the relative errors of Lagrange multipliers and the optimal solutions should be monitored. They should decrease to a sufficiently small level
    \begin{eqnarray*}
    \textrm{Algorithm~1}~\left\{\begin{array}{lr} R_\tau^k=\frac{\sum\limits_{i=1}^N p(\xi_i)\|\phi_{\xi_i}^k-\phi_{\xi}^k\|_{L^1}}{\sum\limits_{i=1}^N p(\xi_i)\|\phi_{\xi_i}^0-\phi_{\xi}^0\|_{L^1}} & (38)\vspace{2mm}\\ R_{v_\xi}^k=\frac{\|v_\xi^k-v_\xi^{k-1}\|_{L^1}}{\|v_\xi^{k-1}\|_{L^1}}~\textrm{with}~v_\xi^k= \sum\limits_{i=1}^N p(\xi_i)v^k(\xi_i) & (39) \vspace{2mm}\\ R_{\phi_\xi}^k=\frac{\|\phi_\xi^k-\phi_\xi^{k-1}\|_{L^1}}{\|\phi_\xi^{k-1}\|_{L^1}} & (40) \end{array}\right.
    \end{eqnarray*}
    \begin{eqnarray*}
    \textrm{Algorithm~3}~\left\{\begin{array}{lr} R_\tau^k=\frac{\sum\limits_{h=1}^n\sum\limits_{i=1}^N p(\xi_i)\|\phi_{h(\xi_i)}^k-\phi_{h(\xi)}^k\|_{L^1}}{\sum\limits_{h=1}^n\sum\limits_{i=1}^N p(\xi_i)\|\phi_{h(\xi_i)}^0-\phi_{h(\xi)}^0\|_{L^1}} & (41)\vspace{2mm}\\ R_{v_{h(\xi)}}^k=\frac{\|v_{h(\xi)}^k-v_{h(\xi)}^{k-1}\|_{L^1}}{\|v_{h(\xi)}^{k-1}\|_{L^1}}~\textrm{with}~v_{h(\xi)}^k= \sum\limits_{i=1}^N p(\xi_i)v_h^k(\xi_i) & (42) \vspace{2mm}\\ R_{\phi_{h(\xi)}}^k=\frac{\|\phi_{h(\xi)}^k-\phi_{h(\xi)}^{k-1}\|_{L^1}}{\|\phi_{h(\xi)}^{k-1}\|_{L^1}} & (43) \end{array}\right.
    \end{eqnarray*}
    where, $\|\cdot\|_{L^1}$ denotes the $L^1$ norm on image domain $\Omega$. All components are calculated in pixel wise. If $R^k<l$ ( $l$ is a small enough parameter), the iteration process will be stopped. Note that Equation (39) and (42) can be quite small if the penalty parameters are large. This is due to their explicit dependence on the penalty parameters.

    The relative energy error should also be considered, we can use the following form:
    \begin{eqnarray*}
    \begin{array}{cr}
    \quad\quad\quad\quad\quad\quad\quad\quad\quad\quad R_e^k=\|E^k-E^{k-1}\|/\|E^{k-1}\| & \quad\quad\quad\quad\quad\quad\quad(44)
    \end{array}
    \end{eqnarray*}
    where $E^k=\sum_{i=1:N} p(\xi_i)E^k(\xi_i)$. The computation stops automatically when $R_e^k$ is less than a predefined tolerance, which indicates that the energy approaches to its steady state.
\item \textbf{For sub minimization problems using ADMM-C in Algorithm 2 and Algorithm 4:} The following constraint errors of $(\vec{w}_{\xi_i}-\nabla \phi_{\xi_i})$  and the relative errors of its corresponding Lagrange multipliers in iterations need to be monitored
    \begin{eqnarray*}
    \textrm{Algorithm~2}~\left\{\begin{array}{lr} R_{\vec{w}_{\xi}}^k=\frac{\sum\limits_{i=1}^N p(\xi_i)\|\vec{w}_{\xi_i}^k-\nabla \phi_{\xi_i}^k\|_{L^1}}{\sum\limits_{i=1}^N p(\xi_i)\|\vec{w}_{\xi_i}^0-\nabla \phi_{\xi_i}^0\|_{L^1}} & \quad\quad\quad (45)\vspace{2mm}\\ R_{\vec{\lambda}_\xi}^k=\frac{\|\vec{\lambda}_\xi^k-\vec{\lambda}_\xi^{k-1}\|_{L^1}}{\|\vec{\lambda}_\xi^{k-1}\|_{L^1}}~\textrm{with}~\vec{\lambda}_\xi^k= \sum\limits_{i=1}^N p(\xi_i)\vec{\lambda}^k(\xi_i) & \quad\quad\quad (46)
    \end{array}\right.
    \end{eqnarray*}
    \begin{eqnarray*}
    \textrm{Algorithm~4}~\left\{\begin{array}{lr} R_{\vec{w}_{h(\xi)}}^k=\frac{\sum\limits_{h=1}^n\sum\limits_{i=1}^N p(\xi_i)\|\vec{w}_{h(\xi_i)}^k-\nabla \phi_{h(\xi_i)}^k\|_{L^1}}{\sum\limits_{h=1}^n\sum\limits_{i=1}^N p(\xi_i)\|\vec{w}_{h(\xi_i)}^0-\nabla \phi_{h(\xi_i)}^0\|_{L^1}} & (47)\vspace{2mm}\\ R_{\vec{\lambda}_{h(\xi)}}^k=\frac{\|\vec{\lambda}_{h(\xi)}^k-\vec{\lambda}_{h(\xi)}^{k-1}\|_{L^1}}{\|\vec{\lambda}_\xi^{k-1}\|_{L^1}}~\textrm{with}~\vec{\lambda}_{h(\xi)}^k= \sum\limits_{i=1}^N p(\xi_i)\vec{\lambda}_h^k(\xi_i) & (48)
    \end{array}\right.
    \end{eqnarray*}
    All numerical quantities are presented in log scale. Some specific methods are used to tune parameters in the implementation process of the proposed approach. The two parameters in $g(\kappa)=\alpha+\beta |\kappa |$, $\alpha$ and $\beta$, control the length and curvature of the segmentation boundary. The ratio between a and b is related to the connectivity and smoothness of the level lines. As discussed in \cite{ztch2013}, the connection of disconnected level lines and smoothness of level lines can be guaranteed by a large parameter $\beta$. In addition, how to determine another two parameters: $\tau$ and $\mu$ associated with Lagrange multipliers will be illustrated. Numerical indicators give the basis of penalty parameter adjustment. One example of their value selection is given in Experiment 4.1.
\end{itemize}

\section{Experimental results}

We apply the proposed segmentation formulations and developed algorithm extensively on various synthetic and real images for multiple purposes. Experimental results are used to validate the performance and efficiency of our proposed models and algorithm. All the experiments are implemented on the same platform (Matlab 8.2) on a PC (Intel (R), CPU: 2.80GHz, RAM: 16GB, cores number: 4, architecture: 64-bit).

\subsection{Experiments for two phase cases on Synthetic Images}

Some Synthetic images of size $256\times256$ pixels are used as the test images. In these experiments, two-phase CV model [2] and the CVE model [11] are used for comparison in order to show the performance of our proposed model. First, we set $v^0(\xi_i)=0$, $\vec{w}^0(\xi_i)=\vec{0}$, $\vec{\lambda}^0(\xi_i)=\vec{0}$ and all the Lagrange multipliers are initially set to be 0 for all the numerical experiments in this section. The same initialization of variables in each experiment are used in order to have a relatively fair comparison. In Figure 1, some results of the CV model, CVE model and our proposed model are first presented respectively. The original images, noisy images with stochastic noises including the Gaussian noise, Rayleigh noise, Poisson noise and Gamma noise, and initialization for $\phi^0$ are shown in (a) and (e). In addition, the pepper $\&$ salt noise is additionally contained in (e). And results obtained by the CV model are presented in (b) and (f). (c) and (g) give the final results obtained by CVE model, results from our model (6) are presented in (d) and (h) separately. From left to right in (b)-(d) and (f)-(h), we start with the optimal solution $\phi_\xi$, followed by final curves plotted on noisy images (red lines) and final curves plotted individually (blue lines). It helps to distinguish the detailed differences among results obtained from different models by presenting the final results in blue separately. It can be clearly seen that the results obtained by our model (6) are much better than those two models. The results obtained from CV model are totally different since it is driven by the mean level of the target region resulting in the fact that it can not recognize whether one particular pixel belongs to big noises or objects. The CVE model's performance is unsatisfactory when the desired object has similar density as the background. With the increase of the homogeneity degree, this kind of drawback becomes more obvious. In this experiment, the parameters for CV model, CVE model and our proposed model are given as follows
\begin{table}[H]
\centering \tabcolsep 10pt
\begin{tabular}{cc|cc}
\hline
\multicolumn{2}{c|}{CV model} & \multicolumn{2}{c}{CVE model}\\
\hline
\multirow{2}{*}{Figure 1 (b):} & $\mu=20,\gamma=3$ & \multirow{2}{*}{Figure 1 (c):} & $\alpha=3,\beta=15,\mu=20$ \\
& $\alpha_1=10,\alpha_2=10$ & & $\alpha_1=8,\alpha_2=8$\\
\hline
\multirow{2}{*}{Figure 1 (f):} & $\mu=20,\gamma=3$ & \multirow{2}{*}{Figure 1 (g):} & $\alpha=3,\beta=10,\mu=20$ \\
& $\alpha_1=7,\alpha_2=7$ & & $\alpha_1=8,\alpha_2=8$ \\
\hline
\multicolumn{4}{c}{Our proposed model (6) via PHA with ADMM-C}\\
\hline
\multirow{2}{*}{Figure 1 (d):} & \multicolumn{3}{c}{ $\alpha=3,\beta=25,\tau=5,\mu=20$ }\\
& \multicolumn{3}{c}{$\alpha_1=10,\alpha_2=10,p(\xi)=(0.4,0.1,0.3,0.2)$}\\
\hline
\multirow{2}{*}{Figure 1 (h):} & \multicolumn{3}{c}{ $\alpha=3,\beta=25,\tau=5,\mu=20$ }\\
& \multicolumn{3}{c}{$\alpha_1=7,\alpha_2=7,p(\xi)=(0.4,0.1,0.3,0.2)$}\\
\hline
\end{tabular}
\end{table}

\begin{figure}[H]
\centering
\subfigure[Original, noisy images and initial contour]{\includegraphics[height=2cm]{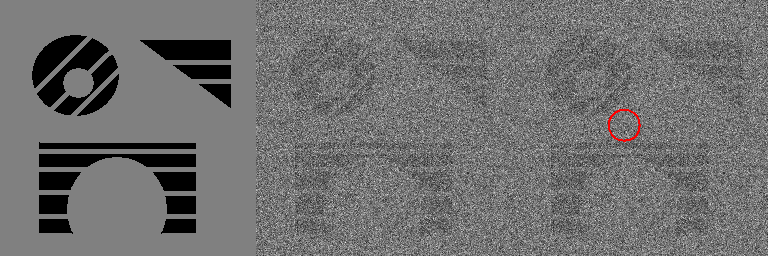}} \qquad\quad\
\subfigure[CV model results]{\includegraphics[height=2cm]{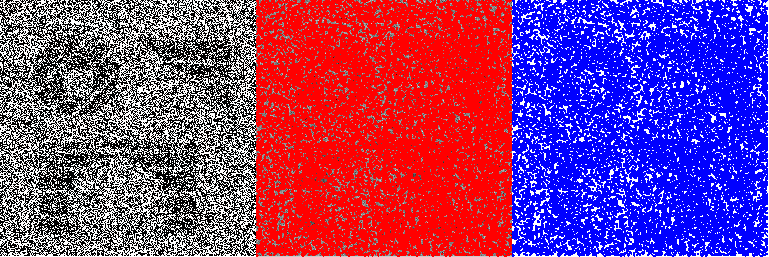}}  \\

\subfigure[CVE model results]{\includegraphics[height=2cm]{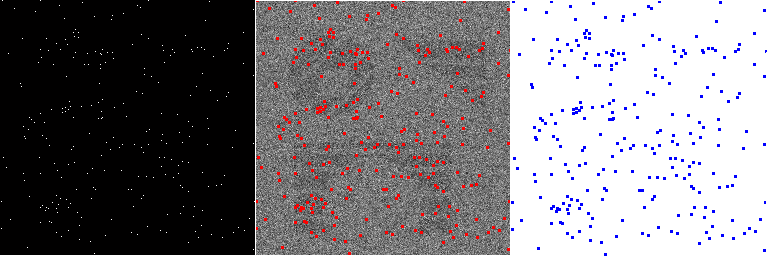}} \quad\qquad
\subfigure[Our model (6) results]{\includegraphics[height=2cm]{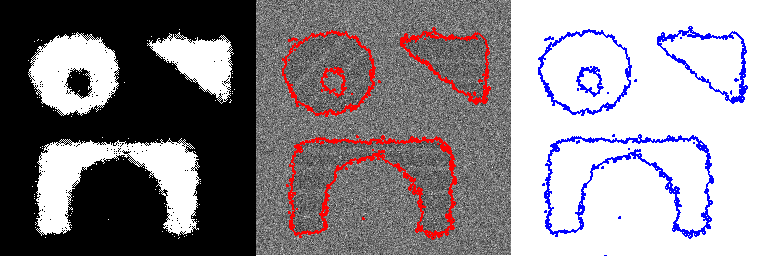}}  \\

\subfigure[Original, noisy images and initial contour]{\includegraphics[height=2cm]{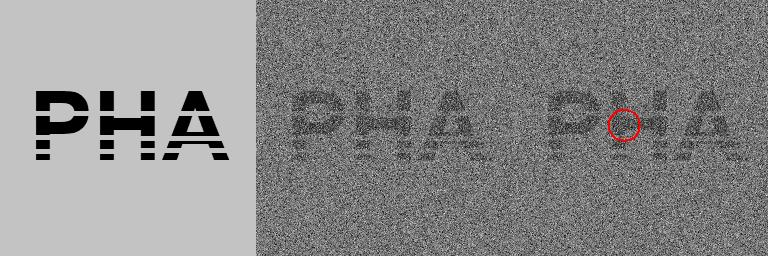}} \qquad\quad\
\subfigure[CV model results]{\includegraphics[height=2cm]{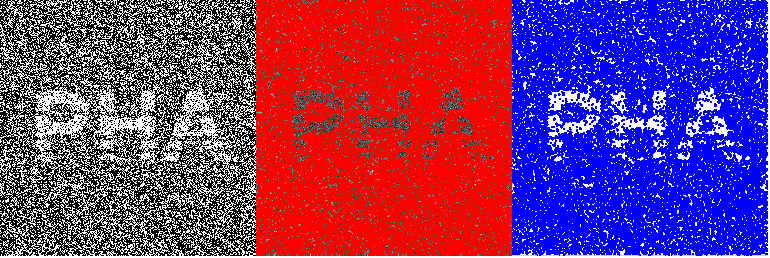}}  \\

\subfigure[CVE model results]{\includegraphics[height=2cm]{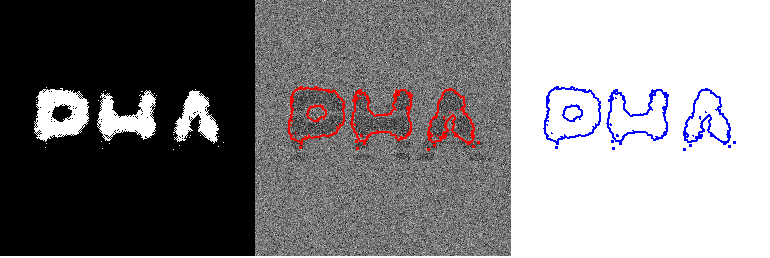}} \qquad\quad\
\subfigure[Our model (6) results]{\includegraphics[height=2cm]{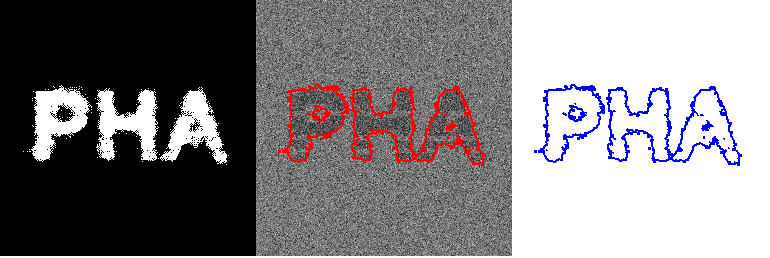}}  \\

\caption{\scriptsize Two phase segmentation for synthetic noisy images with incomplete shapes and letters. (a) and (e): original images, noisy images and the same initial $\phi^0$; (b) and (f): results obtained by CV model; (c) and (g): results obtained by CVE model; (d) and (h): final results from our proposed model (6).}
\end{figure}

Figure 2 gives an example that illustrates the convergence of the relative residuals (Eqn. 38 and 45), the relative errors of Lagrange multipliers (Eqn. 39 and 46), the relative error of $\phi_\xi^k$ (Eqn. 40) and the energy curve (Eqn. 44) in our model respectively. They are obtained for the image in Figure 1 (d). It is clearly shown that the proposed algorithm has converged well before 100 iterations. They also give an important clue on how to choose the penalty parameters t and m. In order to ensure convergence as well as achieving a high speed of convergence, the errors $R_\tau^k$, $R_{\vec{w}}^k$, $R_v^k$ and $R_{\vec{\lambda}}^k$ should converge steadily with nearly the same speed. If $R_\tau^k$, $R_{\vec{w}}^k$ go to zero faster than the others, $\tau$ and $\mu$ can be decreased and vice versa. $R_\tau^k$, $R_{\vec{w}}^k$ will converge to zero with the same speed as the iteration proceeds and the energy will decrease to a steady constant value when $\tau$ and $\mu$ are chosen properly.

\begin{figure}[H]
\centering
\subfigure[]{\includegraphics[height=3.5cm]{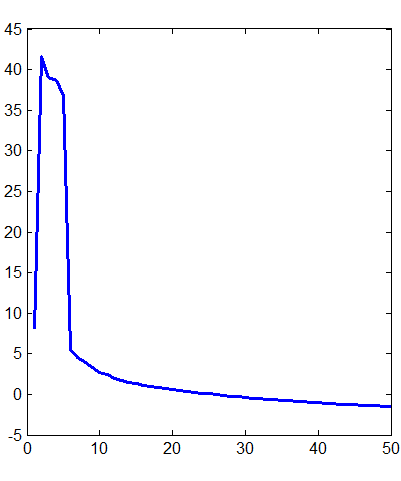}} \qquad
\subfigure[]{\includegraphics[height=3.5cm]{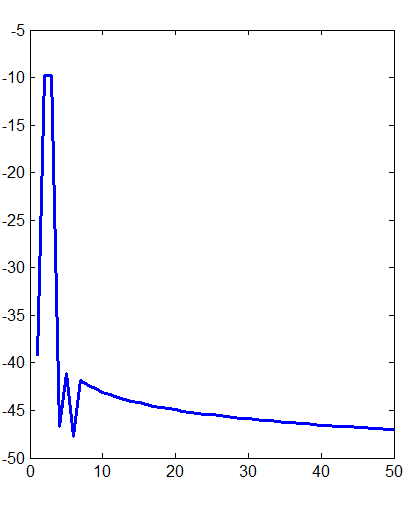}} \qquad
\subfigure[]{\includegraphics[height=3.5cm]{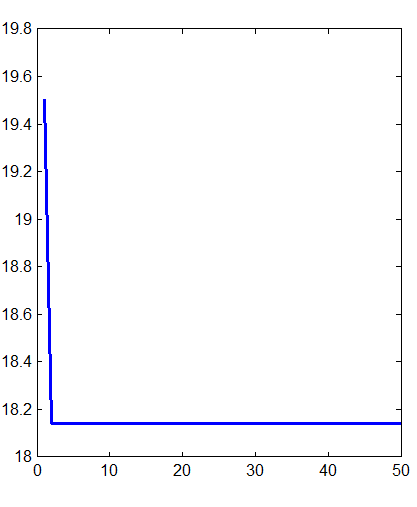}} \qquad
\subfigure[]{\includegraphics[height=3.5cm]{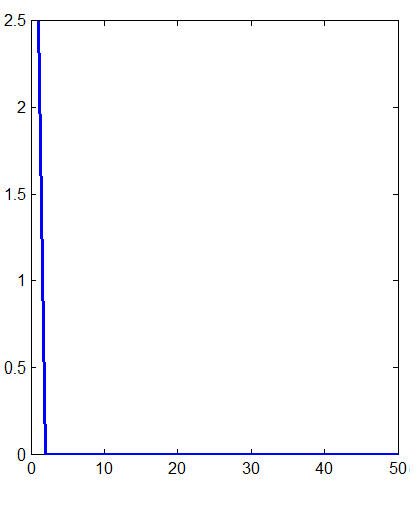}}  \\

\subfigure[]{\includegraphics[height=3.5cm]{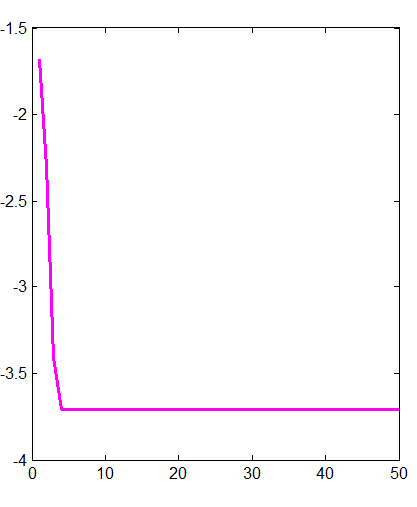}} \qquad
\subfigure[]{\includegraphics[height=3.5cm]{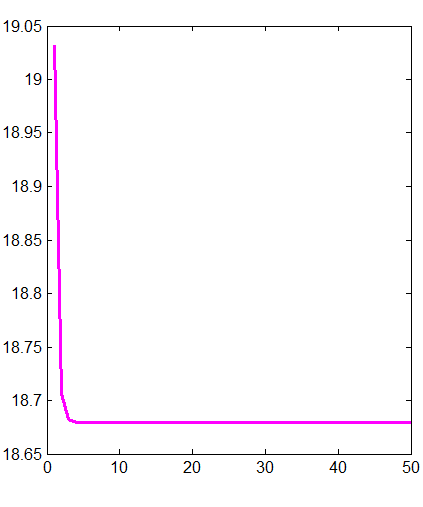}}  \\

\caption{\scriptsize The relative residual plots. (a): auxiliary variables in (38); (b): auxiliary variables in (45); (c): Lagrange multipliers in (39); (d): Lagrange multipliers in (46); (e): optimal solution $\phi_\xi^k$; (f): the energy functional.}
\end{figure}

\subsection{Experiments for two phase cases on real Images}

Real tiger image of size $481\times321$ and butterfly image of size $230\times137$ are used as the test images. In this section, we make visual comparisons with the results subsequently and then show the evolution process of our model. In Figure 3, some results of the color CV model \cite{chs2000}, color CVE model and our proposed model \reff{7} are shown respectively. The color CVE model is stated as
\begin{eqnarray*}
E(\phi,c)=\alpha_1 \int_\Omega \sum_{l=1}^m (f_l-c_{1l})^2\phi dx + \alpha_2 \int_\Omega \sum_{l=1}^m (f_l-c_{2l})^2(1-\phi) dx+ \int_{\Omega} (\alpha+\beta\kappa^2)|\nabla \phi|dx. (49)
\end{eqnarray*}
Figure 3 (a), (g) show the real noisy images. (b), (h) give the initialization of $\phi^0$. The segmented images by using the color CV model are shown in (c) and (i). Figure 3 (d) and (j) show the results of color CVE model. Our proposed model's performance is reflected in (e) and (k). (f) and (l) show the intermediate evolution process of the contour $\phi_\xi$ obtained from the proposed model. In this experiment, we directly choose the results from color CV model as the input of $\phi^0$. One can also initialize $\phi^0$ randomly while the final results vary a little. One feature in both of these two images is that a few discontinuous stripes on the tiger's tail or sparse spots on the butterfly's wings. With the proceeding of iteration, it can be observed that CV model fails to capture the correct boundaries of objects, while both CVE and our model are able to complete an intact shape regardless of the existing gaps within the objects. However, CVE model is inevitably influenced by the stochastic noises especially when these noises smear the elongated structures like the tiger's tail or they increase the homogeneity degree. The parameters used to obtain Figure 3 (c)-(e) and (i)-(k) are
\begin{table}[H]
\centering \tabcolsep 10pt
\begin{tabular}{cc|cc}
\hline
\multicolumn{2}{c|}{Color CV model} & \multicolumn{2}{c}{Color CVE model}\\
\hline
\multirow{2}{*}{Figure 3 (c):} & $\mu=3,\gamma=3$ & \multirow{2}{*}{Figure 3 (d):} & $\alpha=3,\beta=8,\mu=20$ \\
& $\alpha_1=8,\alpha_2=5$ & & $\alpha_1=5,\alpha_2=5$\\
\hline
\multirow{2}{*}{Figure 3 (i):} & $\mu=3,\gamma=3$ & \multirow{2}{*}{Figure 3 (j):} & $\alpha=3,\beta=20,\mu=20$ \\
& $\alpha_1=8,\alpha_2=5$ & & $\alpha_1=7,\alpha_2=7$ \\
\hline
\multicolumn{4}{c}{Our proposed model (6) via PHA with ADMM-C}\\
\hline
\multirow{2}{*}{Figure 3 (e):} & \multicolumn{3}{c}{ $\alpha=3,\beta=4,\tau=5,\mu=80$ }\\
& \multicolumn{3}{c}{$\alpha_1=4,\alpha_2=4,p(\xi)=(0.6,0.1,0.2,0.1)$}\\
\hline
\multirow{2}{*}{Figure 3 (k):} & \multicolumn{3}{c}{ $\alpha=3,\beta=16,\tau=3,\mu=50$ }\\
& \multicolumn{3}{c}{$\alpha_1=4,\alpha_2=4,p(\xi)=(0.6,0.1,0.2,0.1)$}\\
\hline
\end{tabular}
\end{table}

\begin{figure}[H]
\centering
\subfigure[Noisy tiger image]{\includegraphics[height=2.2cm]{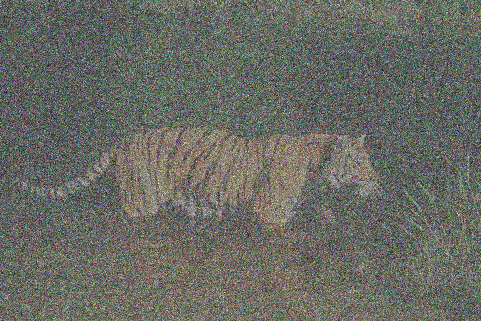}} \qquad\quad
\subfigure[Initialization of $\phi^0$]{\includegraphics[height=2.2cm]{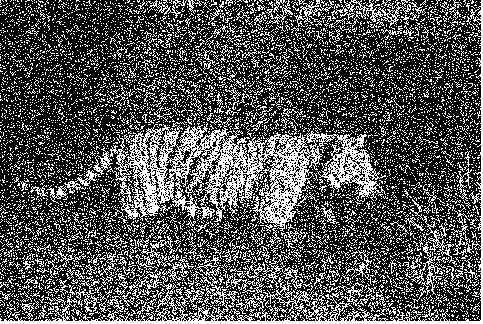}}  \\

\subfigure[Color CV model result]{\includegraphics[height=2.2cm]{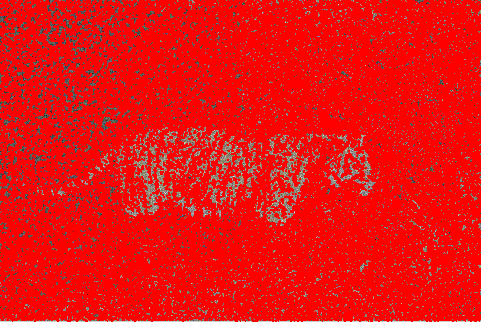}} \qquad
\subfigure[Color CVE model result]{\includegraphics[height=2.2cm]{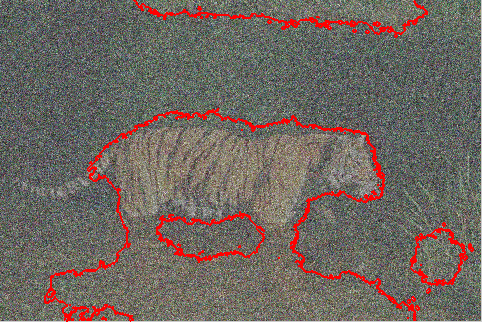}} \qquad
\subfigure[Our model (7) result]{\includegraphics[height=2.2cm]{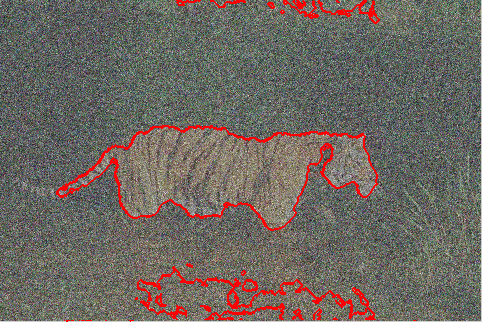}} \\

\subfigure[Intermediate curve evolution from our model (7)]{\includegraphics[height=2.2cm]{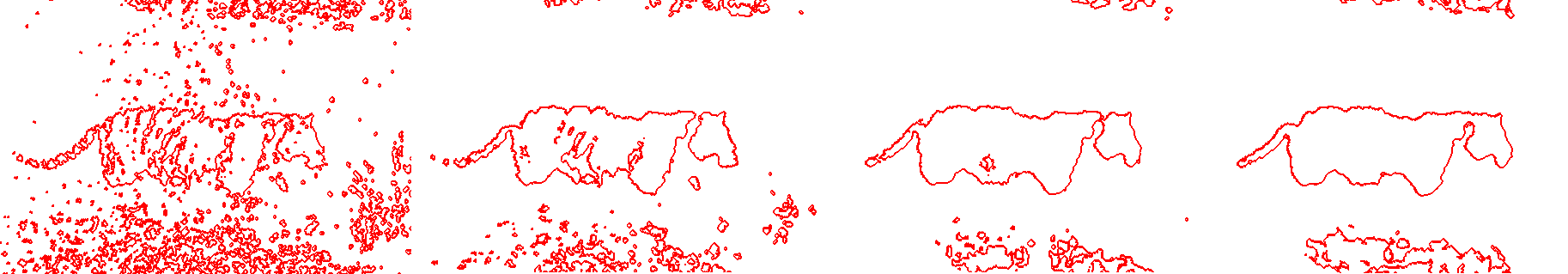}} \\

\subfigure[Noisy tiger image]{\includegraphics[height=2.2cm]{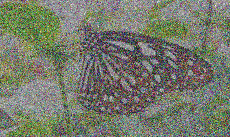}} \qquad\quad
\subfigure[Initialization of $\phi^0$]{\includegraphics[height=2.2cm]{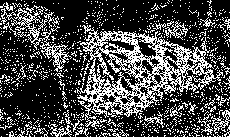}}  \\

\subfigure[Color CV model result]{\includegraphics[height=2.2cm]{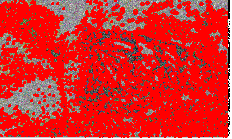}} \qquad
\subfigure[Color CVE model result]{\includegraphics[height=2.2cm]{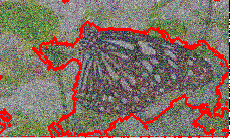}} \qquad
\subfigure[Our model (7) result]{\includegraphics[height=2.2cm]{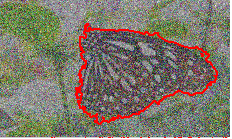}} \\

\subfigure[Intermediate curve evolution from our model (7)]{\includegraphics[height=2cm]{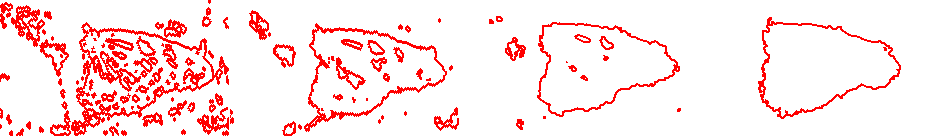}}

\caption{\scriptsize Two phase segmentation for real noisy images with incomplete shapes. (a) and (g): noisy images; (b) and (h): initial $\phi^0$; (c) and (i): results obtained by CV model; (d) and (j): results obtained by CVE model; (e) and (k): final results from our proposed model (7); (f) and (l): intermediate curve evolution by our proposed model.}
\end{figure}

\subsection{Experiments for segmentation with depth cases on Synthetic Images}

In this section, we will apply the proposed models \reff{23} and \reff{24} using PHA with ADMM-C algorithm into synthetic images compared with classic segmentation with depth model \reff{2} without stochastic programming. The detailed implementation framework is shown as follows and all experiments for segmentation with depth applications follow the same procedure.
\begin{figure}[H]
\centering
\includegraphics[height=3cm]{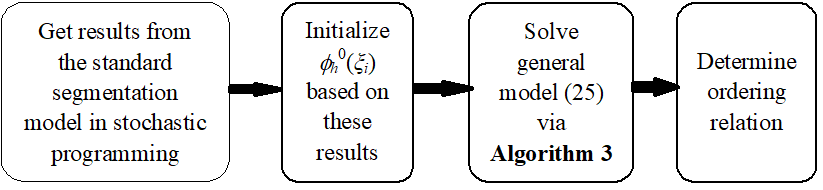}
\end{figure}

In order to speed up the evolution of contours and improve the accuracy of results, we initialize the binary level-set functions $\phi_h^0(\xi_1)=\cdots=\phi_h^0(\xi_N)$ using the results from the the standard multiphase segmentation model in stochastic programming, which is given by
\begin{eqnarray*}\label{50}
&&\arg\min_{\theta_\xi,\phi_\xi\in\{0,1\}} \;\left\{E_{\scriptsize\textrm{SP-general}}^{\scriptsize\textrm{Multi}}(\theta_{\xi},\phi_{\xi})\right.\\
&=&\sum_{h=1}^n \int_{\Omega} |\nabla\phi_h(\xi)|dx+\sum_{h=1}^n \int_{\Omega} Q_h^*(x,\theta_h(\xi))\chi_h(\xi)dx+ \int_{\Omega} Q_{n+1}^*(x,\theta_{n+1}(\xi))\chi_{n+1}(\xi)dx \nonumber\\
&&\left.+\sum_{h=1}^n \int_{\Omega} (v_h^k(\xi)\cdot\phi_h(\xi)+\frac{\tau}{2}(\phi_h(\xi)-\phi_{h(\xi)}^k)^2)dx\right\} \quad\quad\quad\quad \quad\quad\quad\quad (50)
\end{eqnarray*}
Different form the traditional ones \cite{tpldww2017,zchw2006} that initialized contours by the standard multiphase segmentation model without stochastic programming, we take the situation of unknown noises into consideration. In experiments, we find the initialization of $\phi_h^0(\xi_i)$ will be inevitably influenced when big stochastic noises contained in the original image. Then it may lead to a failure for entire framework to obtain expected results. An example is shown below in Figure 4 to explain this situation clearly. Figure 4 (a) gives the synthetic image (size $100\times100$) with two circles corrupted by noises randomly for testing and standard multiphase segmentation results. (b) shows the initialization of the binary level set functions by using the results of standard segmentation method. And the final results obtained by traditional segmentation with depth model \cite{tpldww2017,zchw2006} are presented in (c). It can be observed that traditional implementation framework will not work under the impact of big noises.

\begin{figure}[H]
\centering
\subfigure[Noisy image and standard multiphase segmentation results]{\includegraphics[height=3cm]{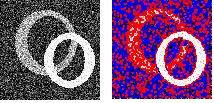}} \qquad\quad
\subfigure[Initial $\phi_h^0$ based on standard multiphase segmentation results]{\includegraphics[height=3cm]{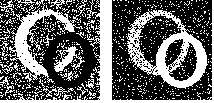}}  \\

\subfigure[traditional segmentation with depth results]{\includegraphics[height=3cm]{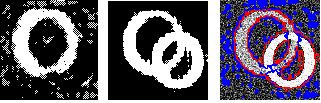}}
\caption{\scriptsize Traditional segmentation with depth for an image with two circles. (a): noisy images and results obtained by the standard multiphase segmentation model; (b): the initialization for two binary functions $\phi_h^0$; (c): results obtained by traditional segmentation with depth model [13,15].}
\end{figure}

Figure 5 presents the novel implementation framework for segmentation with depth in stochastic programming. The same testing image is used in (a). With the application of (50), we obtain separate objects shown in (a) and initialize $\phi_h^0$ in (b). Obvious progress in performance of our proposed model (23) compared with the traditional one lies in (c), the final shapes are reconstructed successfully even though there existing big noises. The parameters used for our proposed model (23) are
\begin{table}[H]
\centering \tabcolsep 10pt
\begin{tabular}{cc}
\hline
\multicolumn{2}{c}{Our proposed model (23) via PHA with ADMM-C}\\
\hline
Figure 5 (c): & $\alpha=3,\beta=10,\tau=5,\mu=30,p(\xi)=(0.5,0.1,0.2,0.2)$\\
\hline
\end{tabular}
\end{table}

\begin{figure}[H]
\centering
\subfigure[Noisy image and results obtained from functional (50)]{\includegraphics[height=3cm]{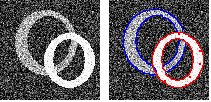}} \qquad\quad
\subfigure[Initial $\phi_h^0$ for two binary functions ]{\includegraphics[height=3cm]{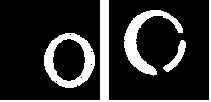}}  \\

\subfigure[Our proposed model (23) results]{\includegraphics[height=3cm]{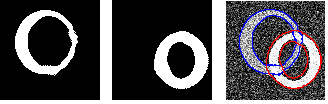}}

\caption{\scriptsize Our proposed model (23) for an image with two circles. (a): noisy images and results obtained by the standard multiphase segmentation in stochastic programming model; (b): the initialization for two binary functions $\phi_h^0$; (c): final results obtained by the proposed model}
\end{figure}

In order to determine the ordering relations of the white circle and gray circle, we minimize the energy functional \reff{23} based on the assumptions that the white circle is occluded by the gray circle or the gray circle is occluded by the white circle. The results are listed in Table 3, from which we can deduce that the white circle, the gray circle and the background are ordered from the nearest to farthest with respect to the observer.

\begin{table}[H]
\centering \tabcolsep 10pt
\begin{tabular}{cc}
\multicolumn{2}{c}{\textbf{TABLE 3} Minimal energies of different ordering relations}\vspace{1mm}\\
\hline
\textbf{Possible Order} & \textbf{Minimum of energy functional}\\
\hline
1. white circle $\Rightarrow$ gray circle & 18.56 \\
2. gray circle $\Rightarrow$ white circle & 20.83 \\
\hline
\end{tabular}
\end{table}

\subsection{Experiments for segmentation with depth cases on real Images}

In the last experiment, one real image with a circle and a hand (size $360\times360$) and the other with a bird and a trunk (size $220\times241$) are shown in Figure 6. Figure 6 (a) and (d) show the original noisy image and the result from standard segmentation model in stochastic programming plotted on the original noisy image. The two initial values for $\phi_h^0$ are given in (b) and (e). The final results from our model \reff{24} are provided in (c) and (f). Our model can clearly perform well in real images. In this experiment, the parameters for our proposed model \reff{24} are given as follows
\begin{table}[H]
\centering \tabcolsep 10pt
\begin{tabular}{cc}
\hline
\multicolumn{2}{c}{Our proposed model (24) via PHA with ADMM-C}\\
\hline
Figure 6 (c): & $\alpha=3,\beta=25,\tau=5,\mu=20,p(\xi)=(0.4,0.1,0.3,0.2)$\\
Figure 6 (f): & $\alpha=3,\beta=25,\tau=3,\mu=10,p(\xi)=(0.5,0.1,0.3,0.1)$\\
\hline
\end{tabular}
\end{table}

\begin{figure}[H]
\centering
\subfigure[Noisy image and results obtained from functional (50)]{\includegraphics[height=2.2cm]{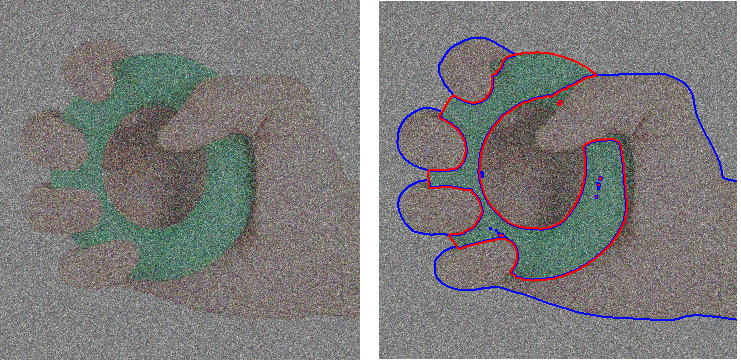}} \qquad\quad
\subfigure[Initial $\phi_h^0$ based on results of functional (50)]{\includegraphics[height=2.2cm]{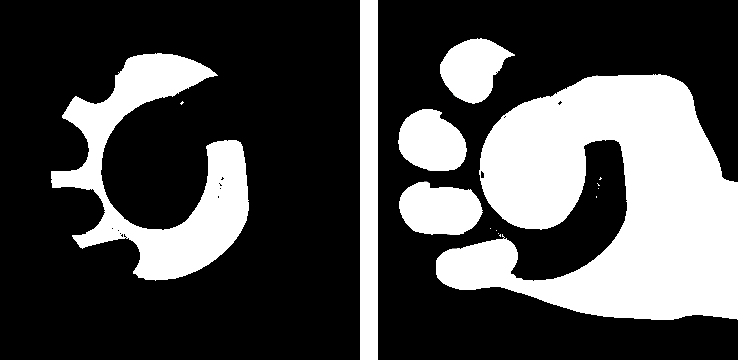}}  \\

\subfigure[Our proposed model (24) results]{\includegraphics[height=2.2cm]{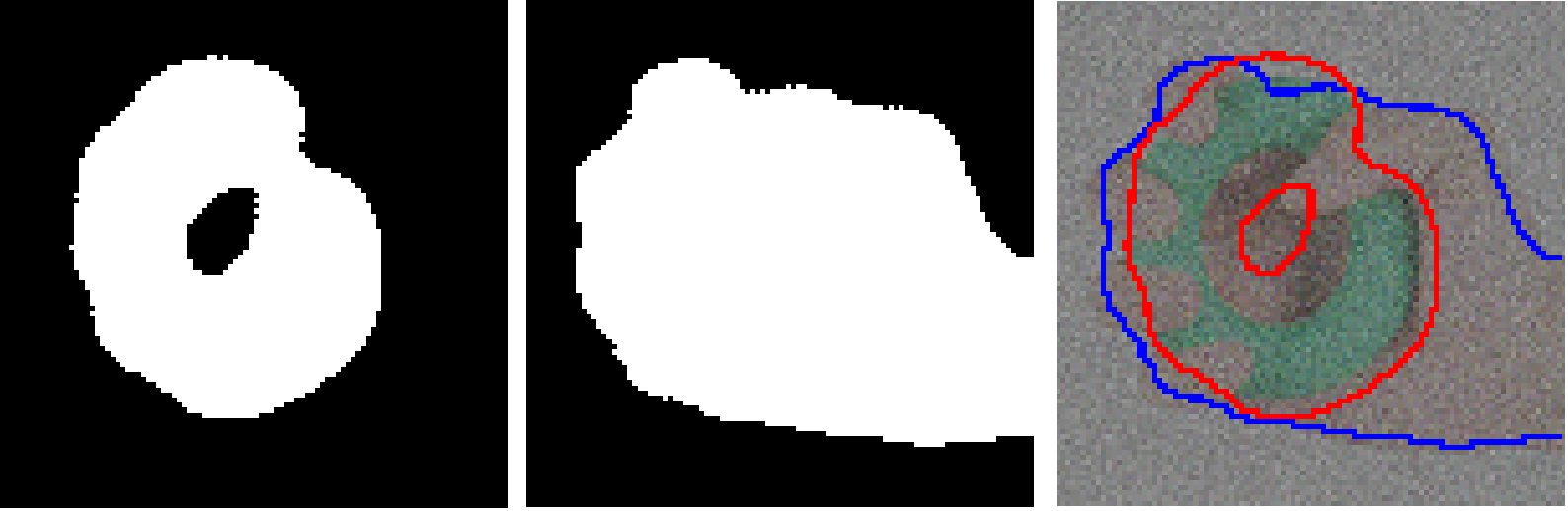}} \\

\subfigure[Noisy image and results obtained from functional (50)]{\includegraphics[height=2.2cm]{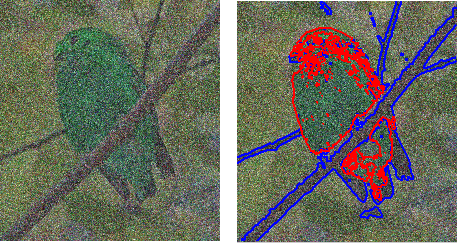}} \qquad\quad
\subfigure[Initial $\phi_h^0$ based on results of functional (50)]{\includegraphics[height=2.2cm]{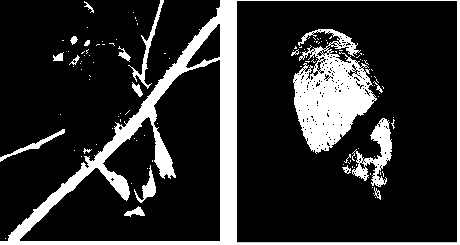}}  \\

\subfigure[Our proposed model (24) results]{\includegraphics[height=2.2cm]{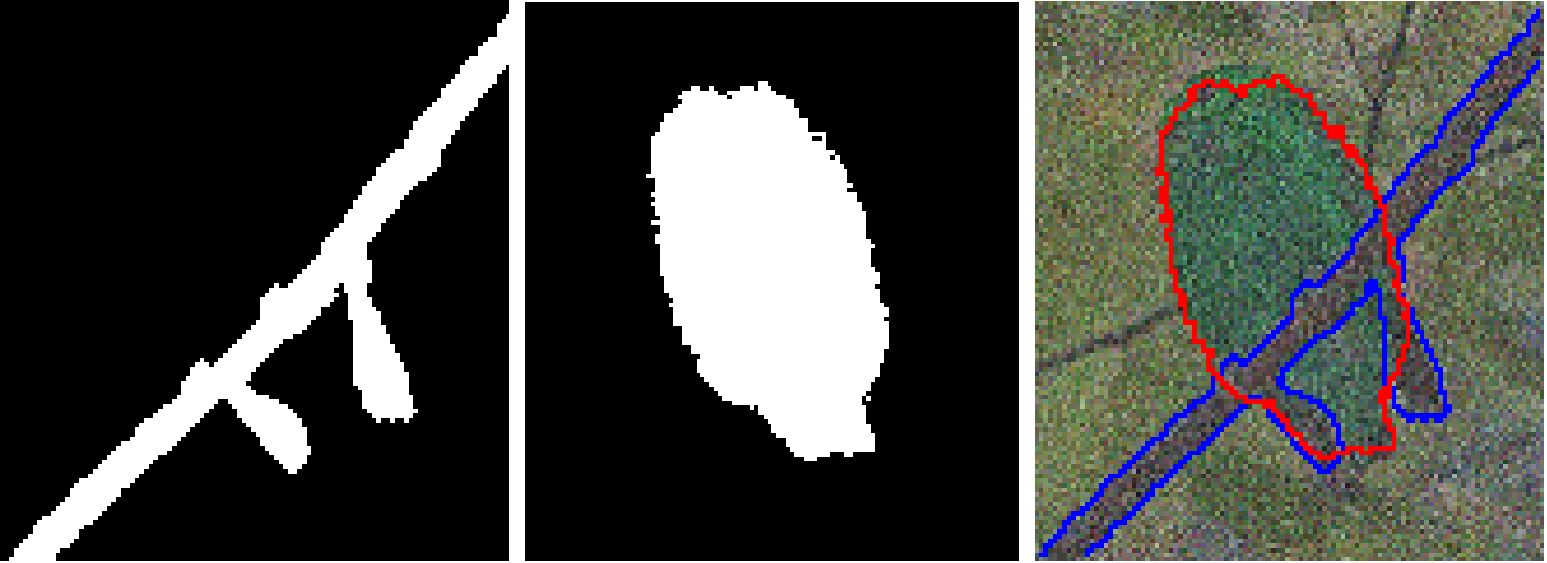}} \\

\caption{\scriptsize Our proposed model (24) for real image. (a) and (d): noisy images and results obtained by the standard multiphase segmentation in stochastic programming model; (b) and (e): the initialization for two binary functions $\phi_h^0$; (c) and (f): final results obtained by the proposed model.}
\end{figure}

For the ordering relations determination for Figure 6 (c) and (f), we minimize the energy functional based on the assumptions including all the potential ordering respectively. From the results listed in Table 4 and Table 5, we can choose the correct orderings which are mapping to the minimal functional values.

\begin{table}[H]
\centering \tabcolsep 10pt
\begin{tabular}{cc}
\multicolumn{2}{c}{\textbf{TABLE 4} Minimal energies of different ordering relations for Figure 6(a) }\vspace{1mm}\\
\hline
\textbf{Possible Order} & \textbf{Minimum of energy functional}\\
\hline
1. green circle $\Rightarrow$ hand & 40.57 \\
2. hand $\Rightarrow$ green circle & 45.26 \\
\hline
\end{tabular}
\end{table}

\begin{table}[H]
\centering \tabcolsep 10pt
\begin{tabular}{cc}
\multicolumn{2}{c}{\textbf{TABLE 5} Minimal energies of different ordering relations for Figure 6(b) }\vspace{1mm}\\
\hline
\textbf{Possible Order} & \textbf{Minimum of energy functional}\\
\hline
1. bird $\Rightarrow$ trunk & 28.56 \\
2. trunk $\Rightarrow$ bird & 20.83 \\
\hline
\end{tabular}
\end{table}

There is one important thing need to be noted. When dealing with the convex optimization problem, we have to use a threshold method to realize the binarization of $\phi_{h(\xi)}^{k+1}$. It is an important way to help find the accurate results. The histograms of non-threshold and threshold results from Figure 6 (c) are given in Figure 7. It gives us a good way to choose the threshold of $\phi_{h(\xi)}^{k+1}$. In practice, we find the threshold $\eta=0.5$ could be applicable.

\begin{figure}[H]
\centering
\subfigure[$\phi_{1\xi}$ without threshold]{\includegraphics[height=4cm]{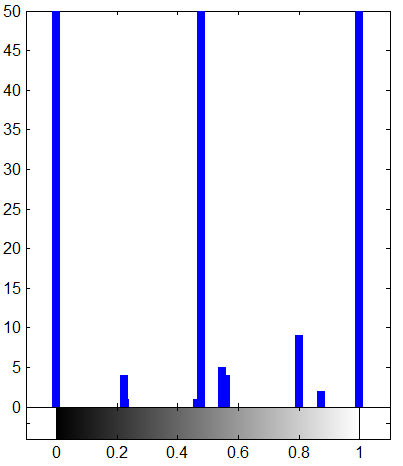}}
\subfigure[$\phi_{2\xi}$ without threshold]{\includegraphics[height=4cm]{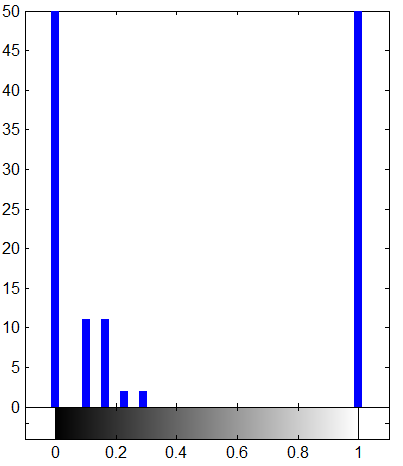}}
\subfigure[$\phi_{1\xi}$ with threshold]{\includegraphics[height=4cm]{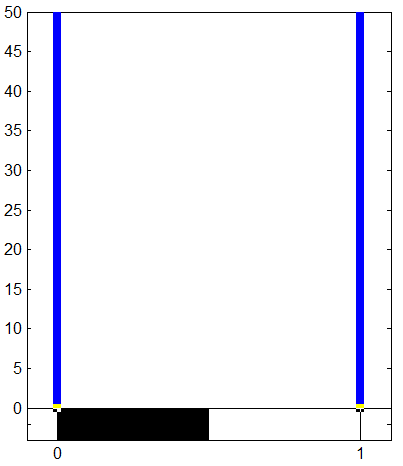}}
\subfigure[$\phi_{2\xi}$ with threshold]{\includegraphics[height=4cm]{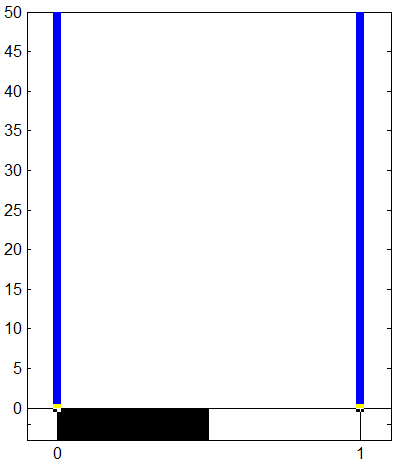}}
\vspace{-2mm}
\caption{Histograms of the final binary level set functions (non-threshold and threshold). (a) and (b): histogram of non-threshold solutions; (c) and (d): histogram of threshold solutions; (a)-(d) are from Figure 6 (c).}
\end{figure}

At last, the efficiency of our proposed PHA with ADMM-C algorithm is emphasized by presenting the number of iterations and computation time in Table 6. The iterations and time are shown according to all of our proposed models applied in Experiment 1 to 4. The computational time is measured in seconds.

\begin{table}[htbp]
\centering \tabcolsep 10pt
\begin{tabular}{ |c|c|c|c|}
\multicolumn{4}{c}{\textbf{TABLE 6} Number of iterations and computational time}\vspace{1mm}\\
\hline
\textbf{Image} & \textbf{Size} & \textbf{Iterations} & \textbf{Time}\\
\hline
Figure 1 (d)    &	$256\times 256$  &	80  &	1.53\\
\hline
Figure 1 (h)    &	$256\times 256$  &	73  &	1.25\\
\hline
Figure 3 (e)    &	$481\times 321$  &	55  &	7.6\\
\hline
Figure 3 (k)    &	$230\times 137$  &	50  &	0.79\\
\hline
Figure 5 (c)    &	$100\times 100$  &	45  &	1.23\\
\hline
Figure 6 (c)    &	$360\times 360$  &	60  &	8.9\\
\hline
Figure 6 (f)    &	$220\times 241$  &	55  &	2.69\\
\hline
\end{tabular}
\end{table}

\section{Conclusions}

We propose novel variational approach for image segmentation with stochastic noises and develop a progressive hedging algorithm to solve them. Our approach possesses three outstanding advantages: 1) improving segmentation ability for noisy images without the prerequisite that one given model and one specific noise distribution are counterpoints; 2) realizing completion of meaningful missing boundaries and reconstruction of occluded structures of objects in a highly noisy background; 3) We incorporate the ADMM method and a curvature weighted approach into the calculation procedure to guarantee the segmentation quality on both convergence and efficiency. Extensive experiments were conducted on images with multiple segmentation purposes which is more challenging due to the limited image quality. Experiment results demonstrate the significant performance improvements of our work. Furthermore, for cases with big stochastic noises and damages, our proposed model achieves better performance than the traditional model, which is of great significance for image understanding with problems such as occlusion, large damages or noises, etc.

In future, our work will focus on embedding other powerful techniques such as deep network with generative capacity into a variational framework to cope with more complicated situations. For instance, segmenting salient objects from images with complex background or lower resolution, even though with clutter and partial occlusions.


\begin{thebibliography}{1}

\bibitem{ms1989}
D. Mumford, J. Shah, Optimal approximations by piecewise smooth functions and associated variational problems. Communications on pure and applied mathematics, 42(5): 577-685, 1989.

\bibitem{ch2001}
T. F. Chan, L. A. Vese, Active contours without edges. Image processing, IEEE transactions on, 10(2): 266-277, 2001.

\bibitem{f2010}
F. Li et al., A multiphase image segmentation method based on fuzzy region competition. SIAM Journal on Imaging Sciences, 3(3): 277-299, 2010.

\bibitem{ywf2012}
H. Yu, W. W. Wang, and X. C. Feng, A new fast multiphase image segmentation algorithm based on nonconvex regularizer. Pattern Recognition, 45(1): 363-372, 2012.

\bibitem{brd2010}
T. Brox, M. Rousson, R. Deriche et al., Colour, texture, and motion in level set based segmentation and tracking. Image and Vision Computing, 28(3): 376-390, 2010.

\bibitem{wyg2018}
B. Wang, X. Yuan, X. Gao et al., A Hybrid Level Set With Semantic Shape Constraint for Object Segmentation. IEEE Transactions on Cybernetics, 2018.

\bibitem{vch2002}
L. A. Vese, T. F. Chan, A multiphase level set framework for image segmentation using the Mumford and Shah model. International journal of computer vision, 50(3): 271-293, 2002.

\bibitem{pd2002}
N. Paragios, R. Deriche, Geodesic active regions: A new framework to deal with frame partition problems in computer vision. Journal of Visual Communication and Image Representation, 13(1-2): 249-268, 2002.

\bibitem{mrgg2004}
P. Martin, P. R$\acute{e}$fr$\acute{e}$gier, F. Goudail, F. Gu$\acute{e}$rault, Influence of the noise model on level set active contour segmentation. IEEE Trans. Pattern Anal. Mach. Intell. 26: 799–803, 2004.

\bibitem{s2013}
A. Sawatzky et al., A variational framework for region-based segmentation incorporating physical noise models. Journal of Mathematical Imaging and Vision 47(3): 179-209, 2013.

\bibitem{ztch2013}
W. Zhu, X. C. Tai, and T. F. Chan, Image segmentation using eulers elastica as the regularization. Journal of scientific computing, 57(2):414-438, 2013.

\bibitem{td2017}
X. C. Tai, and J. M. Duan. A simple fast algorithm for minimization of the elasitica energy combining binary and level set representations. International journal of numerical analysis and modeling, 14(6): 809-821, 2017.

\bibitem{tpldww2017}
L. Tan, Z. Pan, W. Liu, J. Duan, W. Wei, G. Wang: Image Segmentation with Depth Information via Simplified Variational Level Set Formulation. Journal of Mathematical Imaging and Vision, (5): 1-17, 2017.

\bibitem{nms1993}
M. Nitzberg, D. Mumford, and T. Shiota, Filtering, Segmentation, and Depth, Lecture Notesin Computer Sciences, 662, Springer-Verlag, Berlin, 1993.

\bibitem{zchw2006}
W. Zhu, T. F. Chan and S. Esedoglu, Segmentation with Depth: A Level Set Approach, SIAM Journal on Scientific Computing, 28(5):1957-1973, 2006.

\bibitem{kzs2014}
S. H. Kang, W. Zhu, J. H. Shen, Illusory shapes via corner fusion. SIAM Journal on Imaging Sciences, 7(4):1907–1936, 2014.

\bibitem{tll2018}
L. Tan, W. Liu, L. Li et al. A fast computational approach for illusory contour reconstruction, Multimedia Tools and Applications, 1-24, 2018.

\bibitem{ztch2013-1}
W. Zhu, X. C. Tai, and T. F. Chan, Augmented Lagrangian method for a mean curvature based image denoising model. Inverse problems and imaging, 7(4): 1409-1432, 2013.

\bibitem{tlp2018}
L. Tan, W. Liu, Z. Pan, Color image restoration and Inpainting via Multi-Channel Total curvature. Applied Mathematical Modelling, 61:280–299, 2018.

\bibitem{yk2016}
M. Yashtini, S. H. Kang. A Fast Relaxed Normal Two Split Method and an Effective Weighted TV Approach for Euler's Elastica Image Inpainting. SIAM Journal on Imaging Sciences, 9(4): 1552-1581, 2016.

\bibitem{ynl2016}
Y. Yan, F. Nie, W. Li et al., Image classification by cross-media active learning with privileged information. IEEE Transactions on Multimedia 18(12):2494–2502, 2016.

\bibitem{rw2017}
R. T. Rockafellar, R. J. B. Wets, Stochastic variational inequalities: single-stage to multistage. Mathematical Programming, 165(1): 331-360, 2017.

\bibitem{rs2018}
R. T. Rockafellar, J. Sun, Solving monotone stochastic variational inequalities and complementarity problems by progressive hedging, Mathematical Programming, 1-19, 2018.

\bibitem{gpt2016}
R. Glowinski, T. W. Pan, X. C. Tai, Some facts about operator-splitting and alternating direction methods. Splitting Methods in Communication, Imaging, Science, and Engineering, Springer, 19-94, 2016.

\bibitem{bst2011}
E. Bae, J. Shi, and X. C. Tai. Graph cuts for curvature based image denoising. IEEE Transactions on Image Processing, 20(5):1199-1210, 2011.

\bibitem{dgt2018}
L. J. Deng, R. Glowinski, X. C. Tai, A New Operator Splitting Method for Euler's Elastica Model. arXiv preprint arXiv:1811.07091, 2018.

\bibitem{k2017}
P. Kellman et al., Classification images reveal that deep learning networks fail to perceive illusory contours. Journal of vision, 17(10):569–569, 2017.

\bibitem{pg2017}
T. Poscoliero, M. Girelli, Electrophysiological Modulation in an Effort to Complete Illusory Figures: Configuration, Illusory Contour and Closure Effects. Brain topography, 1–16, 2017.

\bibitem{chs2000}
T. F. Chan, B. Y. Sandberg, L. A. Vese. Active Contours without Edges for Vector-Valued Images. Journal of Visual Communication $\&$ Image Representation, 11(2):130-141, 2000.

\bibitem{rof1992}
L. Rudin, S. Osher, E. Fatemi, Nonlinear total variation based noise removal algorithms, Physica D: nonlinear phenomena, 60 (1-4): 259-268, 1992.

\bibitem{chen2006}
T. F. Chan, S. Esedoglu, and M. Nikolova. Algorithms for finding global minimizers of image segmentation and denoising models. SIAM journal on applied mathematics, 66(5):1632-1648, 2006.

\bibitem{ybt2010}
J. Yuan, E. Bae, and X.C. Tai. A study on continuous max-flow and min-cut approaches. In CVPR, USA, San Francisco, 2010.

\bibitem{b2014}
E. Bae et al., A fast continuous max-flow approach to non-convex multi-labeling problems. Efficient algorithms for global optimization methods in computer vision. Springer, Berlin, Heidelberg, 134-154, 2014.

\bibitem{mps2014}
J. M. Morel, A. B. Petro, C. Sbert, Screened Poisson equation for image contrast enhancement, Image Processing On Line, 4: 16-29, 2014.

\bibitem{hl2017}
M. Hong, Z. Q. Luo, On the linear convergence of the alternating direction method of multipliers. Mathematical Programming, 162(1-2): 165-199, 2017.

\bibitem{lsg2017}
Q. Liu, X. Shen, Y. Gu, Linearized admm for non-convex non-smooth optimization with convergence analysis. arXiv preprint arXiv:1705.02502, 2017.

\bibitem{mgkr2015}
M. Myllykoski, R. Glowinski, T. Karkkainen, T. Rossi, A new augmented Lagrangian approach for L1-mean curvature image denoising. SIAM Journal of Imaging Sciences, 8(1):95–125, 2015.

\bibitem{m2015}
M. Yashtini, Alternating Direction Method of Multiplier for Euler's Elastica-Based Denoising, Scale Space and Variational Methods in Computer Vision. Springer International Publishing, 690–701, 2015.

\bibitem{btz2017}
E. Bae, X. C. Tai, W. Zhu, Augmented Lagrangian method for an Euler's elastica based segmentation model that promotes convex contours. Inverse Problems and Imaging, 11(1): 1-23, 2017.



\end{thebibliography}
\end{document}